\PassOptionsToPackage{dvipsnames}{xcolor}

\documentclass[11pt,a4paper,review,3p]{elsarticle}

\usepackage{siunitx}                
\usepackage{threeparttable}         
\usepackage{booktabs}               
\usepackage{adjustbox}              
\usepackage[acronym]{glossaries-extra}  
\usepackage{glossary-mcols}             
\usepackage[en-GB]{datetime2}           
\usepackage{mathtools}                  
\usepackage{accents}                    
\usepackage[inline]{enumitem}           
\usepackage[bottom]{footmisc}           
\usepackage{pifont}                     
\usepackage{tikz}                   
\usepackage{graphicx}
\usepackage{subcaption}
\usepackage{transparent,import}     
\usepackage{placeins}               
\usepackage[numbers]{natbib}
\usepackage{doi}                                    
\usepackage[
  textwidth=1.7cm,
  textsize=singlespacetiny,
                    ]{todonotes}    
\setabbreviationstyle[acronym]{long-postshort-user}
\glssetcategoryattribute{acronym}{nohyperfirst}{true}

\glssetwidest{RANSAC}

\makeatletter
\newglossarystyle{mymcolalttree}{%
	\glossarystyle{alttree}%
	{%
		\begin{multicols}{2}%
			\def\@gls@prevlevel{-1}%
			\mbox{}\par
	}%
		{\par\end{multicols}}%
	\renewcommand{\glossaryentryfield}[5]{%
		\ifnum\@gls@prevlevel=0
		\hangindent\glstreeindent
		\else
		\settowidth{\glstreeindent}{\textbf{\@glswidestname\space}}%
		\hangindent\glstreeindent
		\parindent\glstreeindent
		\fi
		\makebox[0pt][r]{\makebox[\glstreeindent][l]{%
				\glsentryitem{##1}\textbf{\glstarget{##1}{##2}}}}%
		\ifx\relax##4\relax
		\else
		(##4)\space
		\fi
		##3\glspostdescription \space ##5\par
		\def\@gls@prevlevel{0}%
	}%
}%
\makeatother

\AtBeginDocument{\sisetup{math-rm=\mathrm, detect-weight=true, binary-units = true, mode=text,range-phrase = --}}
\DeclareSIUnit\degree{deg}
\DeclareSIUnit\usd{USD}
\DeclareSIUnit\rpm{rpm}
\DeclareSIUnit\gauss{G}
\DeclareSIUnit\pixel{px}
\newsavebox{\largestimage}
\DTMusemodule{british}{en-GB}
\hypersetup{
    colorlinks=true
}

\usepackage{lscape,pdflscape}
\usepackage{amsmath,bm,amsfonts,amsthm,eucal}
\usepackage{multirow,booktabs}
\usepackage{graphicx}

\usepackage{tabularx}
\usepackage{url}
\usepackage{makecell}

\usepackage{epsfig}

\usepackage{threeparttable}
\usepackage{tablefootnote}
\usepackage{footnotehyper}

\def\Figref#1{Figure~\ref{#1}}
\def\figref#1{Fig.~\ref{#1}}
\def\Tabref#1{Table~\ref{#1}}

\def\Secref#1{Section~\ref{#1}}
\def\citref#1{Ref.~\citep{#1}}
\def\citrefpl#1{Refs.~\citep{#1}}

\def\acrexplicit#1{\glsxtrlong*{#1} (\glsxtrshort{#1})}
\def\acrconnect#1#2{%
\ifglsused{#1}
    {\glsxtrshort{#1}{#2}\hspace{-0.25em}%
    }
    {\glsxtrshort{#1}{#2} (\glsxtrlong*{#1})\glsunset{#1}%
    }
}

\def\vtheta{{\bm{\theta}}}

\def\vt{{\bm{t}}}

\def\vx{{\bm{x}}}

\def\mR{{\bm{R}}}

\def\mT{{\bm{T}}}

\DeclareMathAlphabet{\mathsfit}{\encodingdefault}{\sfdefault}{m}{sl}
\SetMathAlphabet{\mathsfit}{bold}{\encodingdefault}{\sfdefault}{bx}{n}

\DeclareRobustCommand\undervec[1]{\underaccent{\vec}{#1}}

\def\Fc{\undervec{\bm{\mathcal{F}}}_{c}}

\def\Ft{\undervec{\bm{\mathcal{F}}}_{t}}

\newcommand{\normltwo}{L^2}

\makeglossaries
\makeglossaries

\newacronym{2d}{2D}{two-dimensional}
\newacronym{3d}{3D}{three-dimensional}
\newacronym{sc}{S/C}{spacecraft}

\newacronym{adr}{ADR}{Active Debris Removal}
\newacronym{ai}{AI}{Artificial Intelligence}
\newacronym{ann}{ANN}{Artificial Neural Network}
\newacronym{alhat}{ALHAT}{Autonomous Landing Hazard Avoidance Technology}

\newacronym{bce}{BCE}{Binary Cross-Entropy}

\newacronym{cl}{CL}{Convolutional Layer}
\newacronym{cnn}{CNN}{Convolutional Neural Network}
\newacronym{coco}{COCO}{Common Objects in Context}
\newacronym{cd}{CD}{Crater Detection}
\newacronym{ci}{CI}{Crater Identification}
\newacronym{cro}{CRO}{Candidate for a Regional Object}

\newacronym{deeptam}{DeepTAM}{Deep Tracking and Mapping}
\newacronym{dmlp}{DMLP}{Deep Multi-Layer Perception}
\newacronym{dem}{DEM}{Digital Elevation Map}
\newacronym{dl}{DL}{Deep Learning}
\newacronym{drcnn}{DRCNN}{Deep Recurrent Convolutional Neural Network}
\newacronym{dnn}{DNN}{Deep Neural Network}
\newacronym[longplural={Degrees-of-Freedom}]{dof}{DoF}{Degree-of-Freedom}

\newacronym{ekf}{EKF}{Extended Kalman Filter}
\newacronym{esa}{ESA}{European Space Agency}

\newacronym{fl}{FCL}{Fully Connected Layer}
\newacronym{fpr}{FPR}{False Positive Rate}
\newacronym{fpga}{FPGA}{Field-Programmable Gate Array}

\newacronym{gmm}{GMM}{Gaussian Mixture Modelling}
\newacronym{gnc}{GNC}{Guidance, Navigation and Control}
\newacronym{gpops}{GPOPS II}{General Purpose Optimal Control Software}
\newacronym{gpu}{GPU}{Graphics Processing Unit}

\newacronym{hda}{HDA}{Hazard Detection and Avoidance}
\newacronym{hrnet}{HRNet}{High-Resolution Net}

\newacronym{icp}{ICP}{Iterative Closest Point}
\newacronym{ilsvrc}{ILSVRC}{ImageNet Large Scale Visual Recognition Challenge}

\newacronym{krn}{KRN}{Keypoint Regression Network}
\newacronym{kpec}{KPEC}{Kelvins Pose Estimation Challenge}

\newacronym{lclf}{LCLF}{Lunar-Centred, Lunar-Fixed Coordinates}
\newacronym{log}{LoG}{Laplacian of Gaussian}
\newacronym{lro}{LRO}{Lunar Reconnaissance Orbiter}
\newacronym{lstm}{LSTM}{Long Short-Term Memory}
\newacronym{lvlh}{LVLH}{Local-Vertical, Local-Horizontal}

\newacronym{mnist}{MNIST}{Modified National Institute of Standards and Technology}
\newacronym{ml}{ML}{Machine Learning}
\newacronym{mlp}{MLP}{Multilayer Perceptron}
\newacronym{mse}{MSE}{Mean Square Error}

\newacronym{nasa}{NASA}{National Aeronautics and Space Administration}
\newacronym{nn}{NN}{Neural Network}
\newacronym{nea}{NEA}{Near-Earth Asteroid}
\newacronym{nst}{NST}{Neural Style Transfer}

\newacronym{odn}{ODN}{Object Detection Network}

\newacronym{pds}{PDS}{Planetary Data System}
\newacronym{ppv}{PPV}{Positive Predictive Value}
\newacronym{pycda}{PyCDA}{Python Crater Detection Algorithm}
\newacronym{pnp}{P$n$P}{Perspective-n-Point}

\newacronym{ransac}{RANSAC}{Random Sample Consensus}
\newacronym{rcnn}{R-CNN}{Region-based Convolutional Neural Network}
\newacronym{rgb}{RGB}{Red-Green-Blue}
\newacronym{roi}{RoI}{Region of Interest}
\newacronym{rpn}{RPN}{Region Proposal Network}
\newacronym{rmse}{RMSE}{Root Mean Square Error}
\newacronym{rnn}{RNN}{Recurrent Neural Network}

\newacronym{sift}{SIFT}{Scale Invariant Feature Transform}
\newacronym{slam}{SLAM}{Simultaneous Localisation and Mapping}
\newacronym{sgd}{SGD}{Stochastic Gradient Descent}
\newacronym{speed}{SPEED}{Spacecraft Pose Estimation Dataset}
\newacronym{spn}{SPN}{Spacecraft Pose Network}
\newacronym{ssd6d}{SSD-6D}{Single Shot Detector 6D}
\newacronym{soc}{SoC}{System-on-a-Chip}

\newacronym{tpr}{TPR}{True Positive Rate}
\newacronym{trn}{TRN}{Terrain Relative Navigation}

\newacronym{urso}{URSO}{Unreal Rendered Spacecraft On-Orbit}

\newacronym{vo}{VO}{Visual Odometry}

\newacronym{wac}{WAC}{Wide Angle Camera}

\journal{Acta Astronautica}

\begin{document}

\begin{frontmatter}

\title{Deep Learning-based Spacecraft Relative Navigation Methods: A Survey}

\author{Jianing Song\corref{cor1}\fnref{fn0,fn1}}
\ead{jianing.song@city.ac.uk}
\author{Duarte Rondao\fnref{fn0,fn1}}
\ead{duarte.rondao@city.ac.uk}
\author{Nabil Aouf\fnref{fn2}}
\ead{nabil.aouf@city.ac.uk}

\address{City, University of London, ECV1 0HB London, United Kingdom}

\cortext[cor1]{Corresponding author}
\fntext[fn0]{Equal contribution}
\fntext[fn1]{Postdoctoral Research Fellow, Department of Electrical and Electronic Engineering}
\fntext[fn2]{Professor of Robotics and Autonomous Systems, Department of Electrical and Electronic Engineering}

\date{}

\begin{abstract}
Autonomous spacecraft relative navigation technology has been planned for and applied to many famous space missions. The development of on-board electronics systems has enabled the use of vision-based and LiDAR-based methods to achieve better performances. Meanwhile, deep learning has reached great success in different areas, especially in computer vision, which has also attracted the attention of space researchers. However, spacecraft navigation differs from ground tasks due to high reliability requirements but lack of large datasets. This survey aims to systematically investigate the current deep learning-based autonomous spacecraft relative navigation methods, focusing on concrete orbital applications such as spacecraft rendezvous and landing on small bodies or the Moon.
The fundamental characteristics, primary motivations, and contributions of deep learning-based relative navigation algorithms are first summarised from three perspectives of spacecraft rendezvous, asteroid exploration, and terrain navigation. 
Furthermore, popular visual tracking benchmarks and their respective properties are compared and summarised. Finally, potential applications are discussed, along with expected impediments.
\end{abstract}


\end{frontmatter}

\printglossary[type=\acronymtype,title={List of Acronyms},style=mymcolalttree,nonumberlist]



\FloatBarrier
\section{Introduction}

\begin{figure}[hbt]
	\centering
	\includegraphics[width=0.85\textwidth]{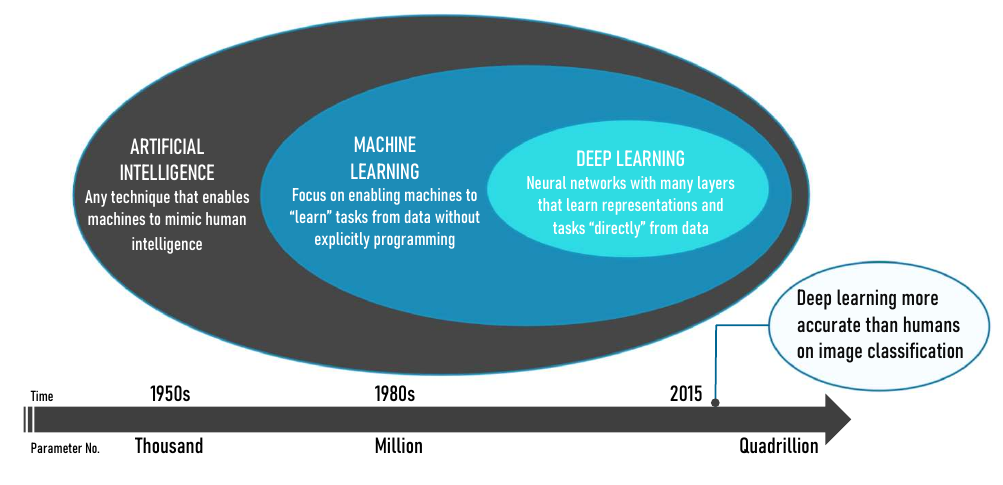}
	\caption{Relationship between \gls{ai}, \gls{ml} and \gls{dl} (reproduced from Mathworks.\protect\footnotemark)}
	\label{fig:mlVSdl}
\end{figure}

\noindent In recent years, there has been a growing interest in \gls{ai}, \gls{ml}, and \gls{dl}, especially amongst science, technology, engineering, and mathematics disciplines. There have been several approaches to define \gls{ai} historically; the most common refers to techniques enabling machines to mimic human intelligence. Then, \gls{ml} is the key component responsible for automatically processing data inside an \gls{ai}. \textcolor{black}{A \gls{nn} is a specific \gls{ml} model aiming to approximate a certain function $f^\ast$ relating training examples $\vx$ to labels $y$ by defining a mapping $y=f(\vx, \vtheta)$ and learning the $\vtheta^\ast$ parameters that result in the best approximation. \glspl{nn} work by stacking many different functions, called layers, and the number of layers defines the depth of the \gls{nn}. The term \gls{dl} derives from this wording, typically signifying a \gls{nn} with large depth \citep{goodfellow2016deep}.} A rough relationship among these three concepts is summarised and illustrated in \figref{fig:mlVSdl}.

\begin{figure}[htb]
	\centering
	\begin{subfigure}[t]{0.32\textwidth}	\centering
    	\includegraphics[height=0.2\textheight]{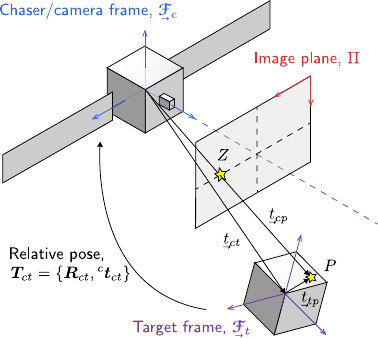}
    	\caption{Rendezvous with non-cooperative spacecraft}
    	\label{fig:intro-scenarios-rv}
	\end{subfigure}
	\begin{subfigure}[t]{0.32\textwidth}	\centering
	    \includegraphics[height=0.2\textheight]{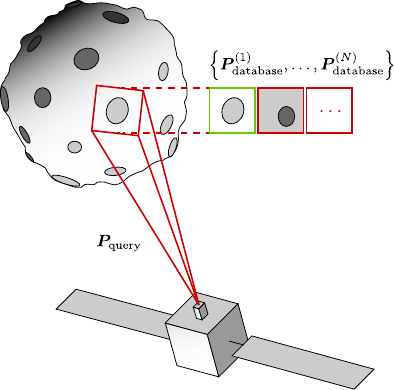}
	    \caption{Asteroid pinpointing via patch classification}
	    \label{fig:intro-scenarios-ap}
	\end{subfigure}
	\begin{subfigure}[t]{0.32\textwidth}	\centering
	    \includegraphics[height=0.2\textheight]{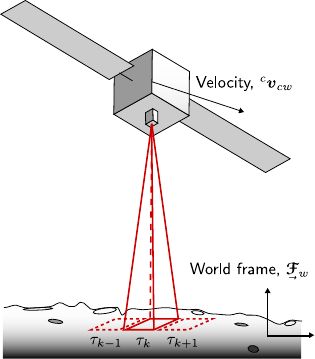}
	    \caption{Terrain navigation}
	    \label{fig:intro-scenarios-trn}
	\end{subfigure}
 \caption{Identification of potential relative navigation scenarios for the application of \Glsxtrshortpl{dnn}.}
\label{fig:intro-scenarios}
\end{figure}

In the field of space exploration, autonomous vision-based \gls{sc} navigation is one key area with the potential of greatly benefiting from \acrconnect{dnn}{-based} estimation methods. Cameras are rapidly becoming the preferred sensor for autonomous rendezvous thanks to the introduction of compact and lightweight passive optical sensors as feasible onboard instruments \citep{wie2014attitude}. Additionally, vision-based techniques have been used in-flight for \textcolor{black}{deep space navigation tasks} \citep{bhaskaran1998orbit}. Potential future applications of domains include:
\begin{enumerate*}[label=\arabic*)]
\item non-cooperative rendezvous with a spacecraft;
\item terrain navigation for descent and landing; and
\item asteroid explorations and asteroid patch pinpoint localisation.
\end{enumerate*}
\footnotetext{\url{https://explore.mathworks.com/machine-learning-vs-deep-learning/chapter-1-129M-833I7.html}.}
All of these scenarios involve the estimation of a chaser or lander spacecraft's relative state, typically through the six \gls{dof} pose $\mT_{ct}$ of the target object frame $\Ft$ relative to the chaser frame $\Fc$, composed of a rotation $\mR_{ct}$, and a translation $\prescript{c}{}{\vt}_{ct}$ (see \figref{fig:intro-scenarios}). Pose estimation methods have traditionally worked by relating features of the target (expressed in $\Ft$), typically obtained from a model, to their images captured by the onboard camera (expressed in $\Fc$), whereas using \gls{dnn} models would adequately capture the intrinsic nonlinearities between the input sensor data and the state estimates, especially for images or \glspl{dem}.

Previous studies have approached the topic of \gls{dl}-based navigation for space. \citet{kothari2020final} collated various applications of \gls{dl} for space, briefly discussing the achieved and prospective goals of onboard systems for spacecraft positioning during docking and landing. 
Aiming at non-cooperative spacecraft rendezvous specifically, \citet{cassinis2019review} first provided a review of \acrconnect{cnn}{-based} schemes in the context of monocular pose estimation systems discussing in detail several works (e.g.\  \cite{shi2018cubesat,sharma2018pose,sharma2019pose}).
\begin{figure}[b!]
     \centering
 	\includegraphics[width=\textwidth]{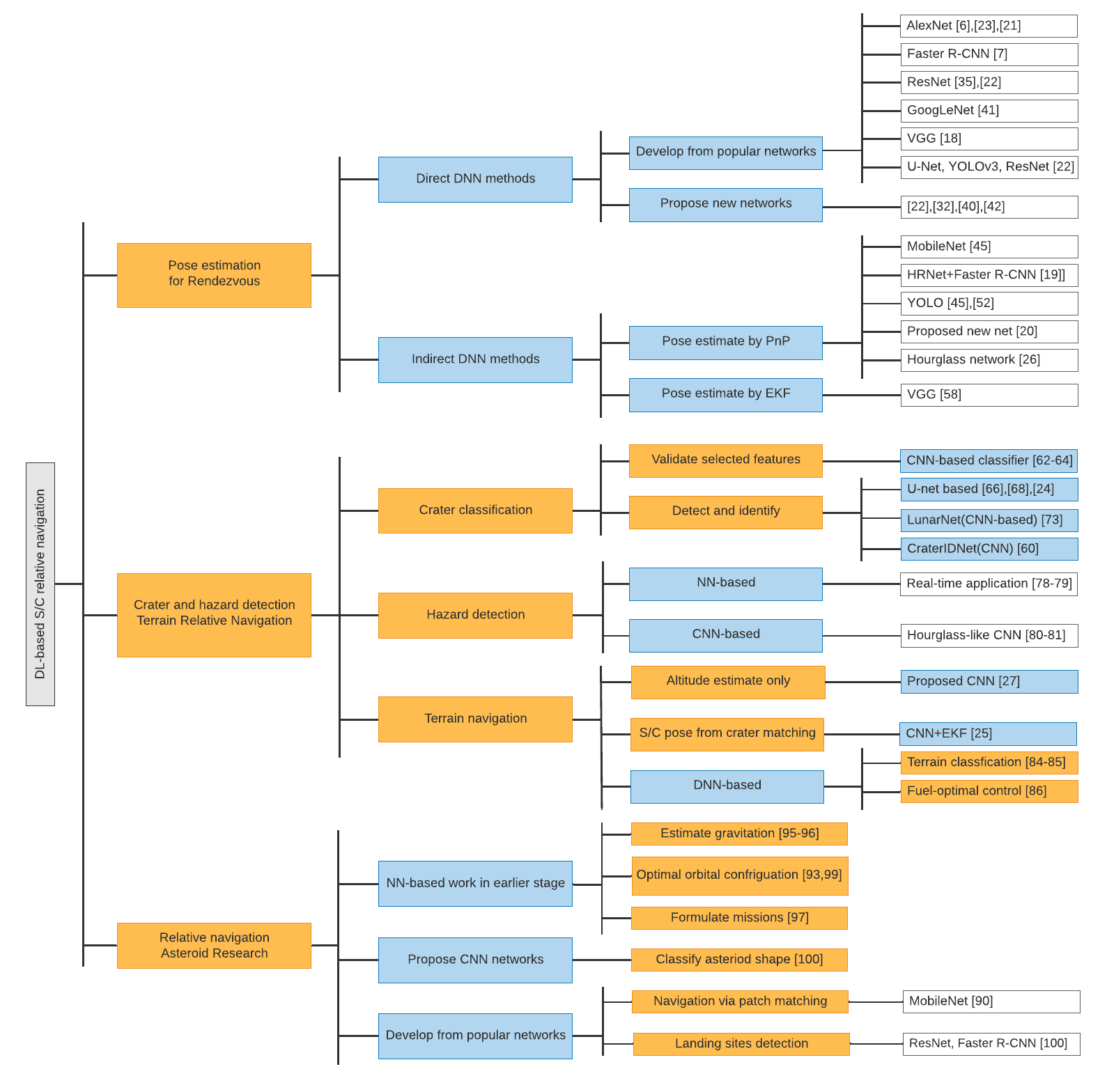}
 	\caption{The tree diagram of \gls{dl}-based \gls{sc} relative navigation approaches reviewed in this paper. \textcolor{black}{The boxes in yellow, blue, and white represent applications, methods, and candidate references, respectively.}}
 	\label{fig:overallview}
\end{figure}
However, there is a shortage of comparative analysis of \gls{dl} methods for general relative navigation in space. With this survey, we thus intend to bridge this gap and provide a comprehensive reference for researchers and engineers aspiring to leverage deep learning for this subject, specifically for the three main applications identified in \figref{fig:intro-scenarios}. 
\figref{fig:overallview} shows the set of research methods and application domains covered by our survey. \textcolor{black}{In \figref{fig:overallview}, direct \gls{dnn} methods are end-to-end methods using \glspl{dnn}, which constitute a direct, uninterrupted pipeline from inputs $x$ to the desired quantity to estimate $y$. In contrast, indirect \gls{dnn} methods are those in which the \gls{dnn} is exclusively tasked with performing the image processing functions on the input, while the actual quantity to be estimated is achieved by combining this output with other methods, such as classical ML, geometry-based optimisation, and Kalman filtering.}


This paper is organized as follows. \Secref{sec:poseestimation} presents a review of \gls{dl}-based pose estimation algorithms for spacecraft rendezvous. \Secref{sec:craterdetection} contains a detailed review of crater and hazard detection of \gls{trn} using \glspl{dnn}. \Secref{sec:asteroid} provides a review of \gls{dl} techniques with a focus on asteroid exploration. Finally, \Secref{sec:conclusion} lists the main conclusions and discussions.

\section{\glsfmtshort{dl}-based Pose Estimation for Spacecraft Relative Navigation }
\label{sec:poseestimation}

\subsection{Related Works on Terrestrial Pose Estimation}

\noindent With the successful application of deep learning approaches in various research areas, \acrconnect{dl}{-based} camera-relative pose determination techniques for terrestrial scenarios have been attracting a considerable amount of interest.

\citet{kendall2015posenet} proposed the PoseNet architecture for 6-\gls{dof} motion estimation in an end-to-end manner. To develop the pose regression network, they used a modified pre-trained GoogLeNet \citep{szegedy2015going} by replacing all softmax classifiers with affine regressor. A weighted sum of the $\normltwo$ error norms of the position vector and the attitude quaternion is selected as the loss function for better training of the location and orientation simultaneously. Their results demonstrate a \SI{2}{m} and \SI{3}{deg} accuracy for large scale outdoor scenes and \SI{0.5}{m} and \SI{5}{deg} accuracy indoors.

Rather than self-localising with respect to a known world model, \citet{wang2017deepvo} presented the DeepVO architecture to obtain a vehicle's egomotion from frame to frame based on monocular \gls{vo}. The pipeline follows the architecture of a \gls{drcnn} \citep{donahue2015long}, in which a pre-trained FlowNet \citep{dosovitskiy2015flownet} first learns features from sequences of \gls{rgb} images, which are then processed by \gls{lstm} cells to estimate poses. The end-to-end \gls{drcnn} framework achieves an average \gls{rmse} drift of \SI{5.96}{\percent} and \SI{6.12}{deg} per trajectory for position and attitude, respectively, on lengths of \SIrange{100}{800}{\metre}, showing a competitive performance relative to Monocular VISO2 \citep{geiger2011stereoscan}.


Differing from the above end-to-end, or direct, methods, some works opt instead by following indirect methods, for which the \gls{dnn} is exclusively tasked with performing the image processing functions on the input, while the actual quantity to be estimated is achieved by combing this output with other methods, such as classical \gls{ml} or Kalman filtering. For instance, \citet{rad2017bb8} developed the BB8 algorithm for object pose estimation by combining a \gls{cnn} to regress the \gls{2d} locations of the eight \gls{3d} points defining their bounding box with a \gls{pnp} algorithm \citep{szeliski2010computer} to retrieve the pose based on those correspondences. The VGG architecture \citep{simonyan2014very} was chosen as the basis for their work, and the classical reprojection (or geometric) error was used as the corresponding loss function \citep{hartley2004multiple}.

\begin{figure}[t]
    \centering
	\includegraphics[width=\textwidth]{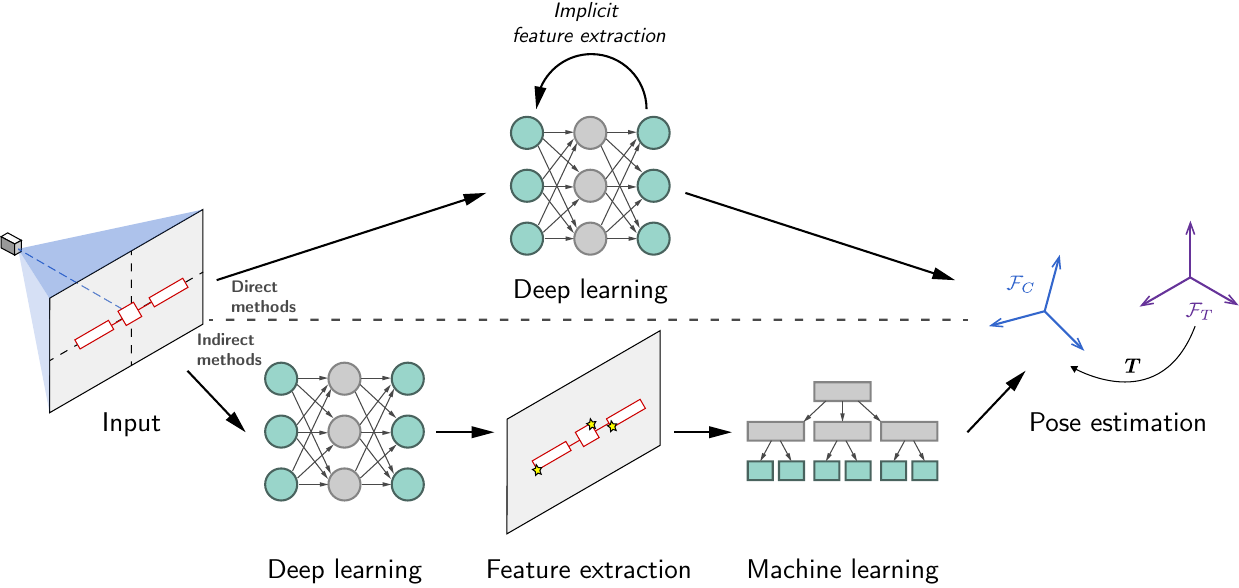}
	\caption{Direct versus indirect methods for \Glsxtrshort{dl}-based pose estimation. The former use a \Glsxtrshort{dnn} to directly estimate the pose from the input, whereas the latter use it exclusively to identify features, or landmarks, on the target, which are then input to a \Glsxtrshort{ml} algorithm.}
	\label{fig:rv-direct-and-indirect}
\end{figure}

\textcolor{black}{\Figref{fig:rv-direct-and-indirect} illustrates the difference between direct and indirect methods, which are explored further in this section.}


\subsection{Challenges and Motivations}

\noindent Recent advancements in \gls{dl} exhibit promising alternatives with respect to classical approaches, and related terrestrial frameworks also inspire the idea of \acrconnect{dl}{-based} spacecraft relative navigation. However, there still exists a gap between the two domains of application. 

Relative pose estimation of objects in space is a different problem from pose determination of objects on Earth due to the vast differences in environment. Additionally, real labelled on-orbit images required for training \gls{dl} algorithms are expensive and hard to obtain, which leads to a lack of space imagery datasets. Challenges in space missions for applying vision-based \gls{dl} methods can be summarised from previous research  \citep{sharma2018pose, sonawani2020assistive,chen2019satellite,HuanLiu-11,oestreich2020onorbit,CosmasKenichi-21, hirano2018deep,downes2020,Lena8} as follows:
\begin{itemize}
    \item Planets and stars acting as background distractors for the navigation system;
    \item Challenging visual conditions due to lack of atmosphere and light diffusion;
    \item Much stronger shadows and varied illumination conditions resulting in extreme image contrast and low signal-to-noise ratio;
    \item Limited properties of space hardware in power consumption and computational resources (e.g.,\ low sensor resolution);
    \item Training datasets  for non-cooperative navigation of spaceborne objects are scarce;
    \item Concerns over the reliability of \gls{dl} technique preventing their practice in the space industry.
\end{itemize}

\noindent The characteristics of space images also challenge conventional vision-based navigation algorithms for spacecraft, while the \gls{dl} technique provides promising solutions and performance that can alleviate these issues. In terms of dynamic lighting, \acrconnect{dnn}{-based}
schemes show increased robustness in attitude initialisation \cite{oestreich2020onorbit,cassinis2020cnn}. {With the deployment of high-performance devices, \acrconnect{cnn}{-based} methods can not only provide a lower computational complexity in pose acquisition process}, but also reduce the need for complicated dynamics models \citep{campbell2017deep}. Additionally, \gls{dnn} pipelines are able to output various information and be combined with navigation filters or other processes \citep{downes2020}.

Motivated by the attractiveness described above, and to overcome current limitations in spacecraft relative pose estimation, \gls{esa} launched the \gls{kpec}\footnote{\url{https://kelvins.esa.int/satellite-pose-estimation-challenge}.}  in 2019, inviting the community to propose and validate new approaches directly from greyscale images acquired by an on-board camera. The data for training and testing in this challenge consisted of Stanford's \gls{speed}, which contains labelled synthetic images of the Tango satellite, and a smaller, real set of images acquired in laboratory using a replica of the target.

\subsection{Direct Frameworks for Spacecraft Relative Pose Estimation}
\label{sec-sub:directDLforPOSE}

\noindent In this survey, direct \acrconnect{dl}{-based} frameworks are defined as those in which the estimation of the desired quantity is entirely relayed to the \gls{dnn}, thus forming a continuous, uninterrupted pipeline from input to output. For spacecraft relative navigation, the problem is posited as estimating the 6-\gls{dof} pose of a target, $\Ft$, in the frame of reference of a chaser, $\Fc$ (as shown in \figref{fig:intro-scenarios-rv}). The target may be non-cooperative, in which case it will not relay any explicit information to the chaser's onboard navigation system, and the relative pose is estimated from acquired images of the target only. 

By partitioning the relative pose space into discrete hypotheses, a classification framework may be established if the target spacecraft has a known model. \citet{sharma2018pose} have proposed a deep \gls{cnn} for relative pose classification of non-cooperative spacecraft. Taking advantage of transfer learning, AlexNet model \citep{krizhevsky2012imagenet} pre-trained on the large ImageNet dataset \citep{deng2009imagenet} is modified by replacing the last few layers to adapt to the space imagery of the Tango spacecraft flown in the Prisma mission \citep{persson2006prisma}. Ten datasets with different added noises are created from synthetic images. The proposed approach demonstrated greater accuracy than a baseline method using classical pose estimation techniques from \gls{2d}-\gls{3d} feature matching but is deemed not fine enough for any application other than a coarse initialisation.

\begin{figure}[t]
  	\centering
  		\includegraphics[width=0.85\textwidth]{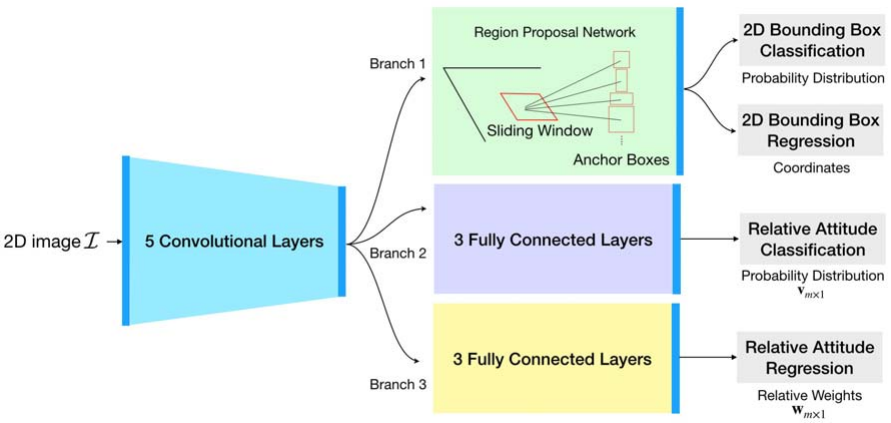}
  \caption{The \acrexplicit{spn} architecture. Reproduced from \citet{sharma2020neural}.}
  \label{fig:supervised-spn}
\end{figure}

\citet{sharma2019pose} later on improve their original work with the creation of the \gls{spn}. The \gls{spn} (\figref{fig:supervised-spn}) uses a five-layer \gls{cnn} backbone of which the activations are connected to three different branches. The first branch uses the Faster \gls{rcnn} architecture \citep{ren2017faster} to detect the \gls{2d} bounding box of the target in the input image. To be robust towards intrusive background elements (i.e., presence of Earth), specific features output by the final activation map of the first branch are extracted using \gls{rcnn}'s \gls{roi} pooling technique, and then fed to the other two branches of the \gls{cnn} containing three fully connected layers.

The second branch classifies the target attitude in terms of a probability distribution of discrete classes. It minimises a standard cross-entropy loss for the $N$ closest attitude labels in the viewsphere. Lastly, the third branch takes the $N$ candidates obtained from the previous branch and minimises another cross-entropy loss to yield the relative weighting of each. The final refined attitude is obtained via quaternion averaging with resort to the computed weights, which can be seen as a soft classification method.

Mathematically, the \gls{spn} utilises a Gauss-Newton algorithm to solve a minimisation problem for the estimate of relative position, for which the required initial guess is obtained from the bounding box (analogously to \citet{kehl2017ssd}). The network is initially trained on the ImageNet dataset, and then the branch layers are further trained with an \SIrange{80}{20}{\percent} train-validation split on the \gls{speed} dataset. As they report, the \gls{spn} method performs at degree-level and centimetre-level on relative attitude and position error, respectively.

In \citref{sharma2020neural}, Sharma and D'Amico expand their conference paper \cite{sharma2019pose} by discussing two features of the \gls{spn}, target-in-target pose estimation and uncertainty quantification.
The capability of estimating the uncertainty associated with the estimated pose of the \gls{spn} emphasises that \gls{spn} can be integrated with conventional navigation filters. Additionally, the authors detail the proposed \gls{speed} dataset, considering the solar illumination of the synthetic images and the ground truth calibration of the relative pose by the real images. The \gls{spn} is also trained in three versions by using different datasets, including \gls{speed}, "Apogee Motor", "Imitation-25", and "PRISMA-25". Experiments are also carried out to demonstrate two key features of \gls{spn} method and compare it with their previous work, namely \acrconnect{cnn}{-based} \citep{sharma2018pose} and image processing-based feature detection and correspondence \citep{Sharma2018robust} methods.   

Instead of employing a bounding box feature detection, \citet{proenca2019deep} modify a pre-trained ResNet architecture \cite{he2016deep} with initial weights trained on the \gls{coco} dataset to keep spatial feature resolution. Similarly to \citref{sharma2019pose}, two branches are designed to estimate \gls{3d} location and orientation, respectively. The position estimation consists of a simple regression branch with two fully connected layers and the relative error is minimised for better generalisation in terms of loss weight magnitudes. The continuous attitude estimation is then realised via a soft classification method \citep{liu2011defense}. Additionally, the authors present their own synthetic \gls{urso} dataset for training featuring Soyuz. Experiments on renders of \gls{urso} and \gls{speed} datasets are conducted to evaluate the proposed framework, with which their model achieved a third and a second place on the synthetic and real test set categories of \gls{speed} in \gls{kpec}, respectively. Moreover, the experimental results show that estimating the orientation by soft classification performs better than direct regression methods. 

\begin{figure}[t]
  	\centering
	\begin{subfigure}[t]{0.7\textwidth}	\centering
		\includegraphics[width=\textwidth]{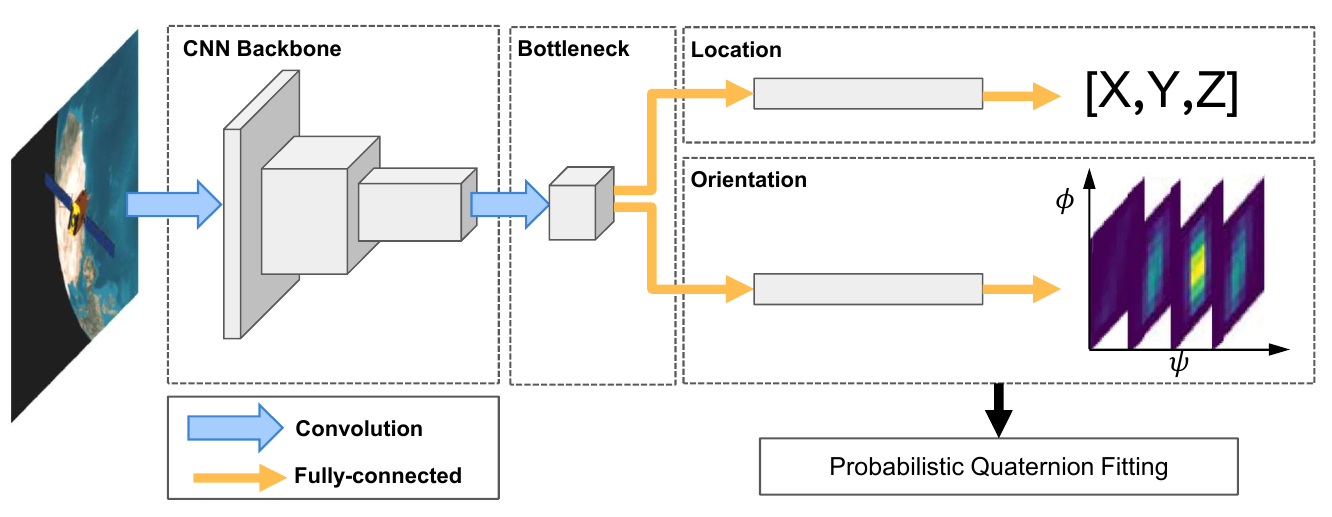}
		\caption{Simplified architecture model}
	\end{subfigure}\hfill
	\begin{subfigure}[t]{0.29\textwidth}	\centering
		\includegraphics[width = \textwidth]{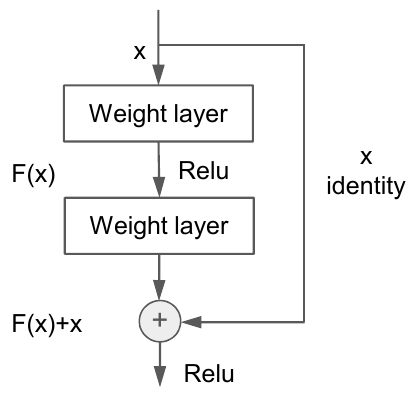}
		\caption{ResNet block}
	\end{subfigure}
\caption{The \gls{cnn} pipeline in \citref{proenca2019deep}. The \gls{cnn} front-end is based on ResNet, of which the elementary blocks implement skip connections that help mitigate the vanishing gradient problem in very deep networks.}
 \label{fig:supervised-proenca}
\end{figure}

\citet{hirano2018deep} present a \gls{3d} keypoint estimator by using an AlexNet-based \gls{cnn} architecture to regress spacecraft pose information directly, rather than retrieving \gls{3d} objects from the location of \gls{2d} keypoints. The parameters of AlexNet are changed for the purpose of the pose estimation task, and batch normalisation layers \citep{ioffe2015batch} are utilised in all \gls{cl} and \gls{fl} for convergence in training. Synthesised images of a \gls{3d} model with the \gls{3d} keypoint position labels are generated on the Gazebo simulator \cite{koenig2004design} and used to train the \gls{cnn}. Real images taken by hardware simulators are imported to evaluate the trained \gls{cnn}. Images in both  training and test dataset include the effects of lighting, shadows, and random noise, which leads the proposed framework to a potential application in practical space missions.

\citet{arakawa2019attitude} also treat the attitude estimation as a \acrconnect{cnn}{-based} regression problem to obtain spacecraft attitude quaternion from the constructed images, in which the output of the proposed \gls{cnn} is four independent real numbers corresponding to four quaternion elements. A \gls{3d} model of the JCSAT-3 satellite is built in the Blender software to generate a training image dataset. A point spread function is applied to the renders for simulating atmospheric fluctuations and optical effects. Compared with conventional image matching approaches, their results clarify an improved performance on the accuracy, robustness, and computational cost.

\begin{figure}[t]
	\centering
	\begin{subfigure}[t]{0.50\textwidth}	\centering
		\includegraphics[width=\textwidth]{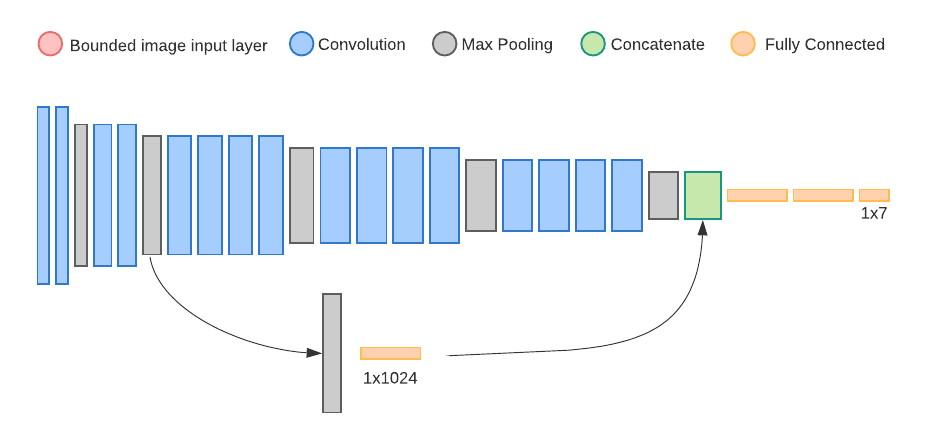}
		\caption{Branch model}
	\end{subfigure}\hfill
	\begin{subfigure}[t]{0.49\textwidth}	\centering
		\includegraphics[width = \textwidth]{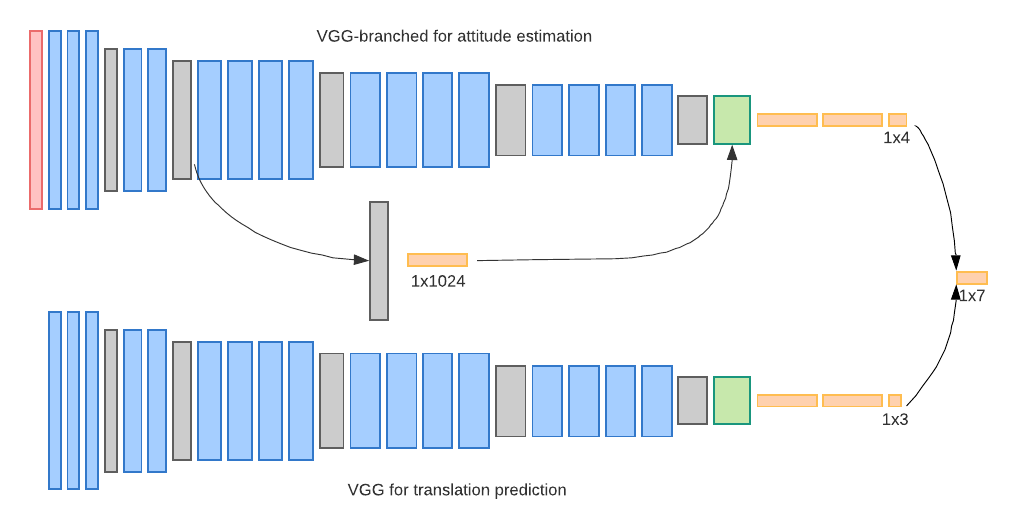}
		\caption{Parallel model}
	\end{subfigure}
	\caption{The VGG-19-based architecture of \citet{sonawani2020assistive}. The branch is used to preserve feature-position information discarded by the later pooling layers.}
	\label{fig:supervised-sonawani2020assistive}
\end{figure}

Considering that natural feature-based methods for spacecraft pose estimation are not always sufficient, \citet{sonawani2020assistive} develop a modified model to assist a cooperative object tracker in space assembly tasks. The proposed \gls{cnn} architecture is similar to \citref{sharma2018pose}, but uses VGG-19 as a backbone and replace the last layer with a 7-node one instead of an activation function. Two different models, namely a branch-based model and a parallel-based model, are developed to estimate relative poses. The frameworks of the two models are illustrated in \figref{fig:supervised-sonawani2020assistive}, in which the parallel model contains two parallel streams for position prediction and attitude estimation, respectively. Synthetic images are generated in Gazebo, including truss-shaped objects labelled with the pose. The Euclidean distance error between the predicted poses and actual ones is defined as the loss function. Simulation results show their models are comparable to the current feature-selection methods and are robust to other types of spacecraft.

Aiming at vision-based uncooperative docking operations, \citet{PhisannupawongKamsing-22} construct a spacecraft pose estimation model by proposing an advanced GoogLeNet pre-trained on \gls{urso}. The original GoogLeNet framework is modified by using 23 layers of \gls{cnn} presented in \citref{kendall2015posenet}, and the output for spacecraft pose is a seven-element vector. Experiments are carried out with an exponential loss function and a weighted Euclidean loss function, separately. The simulating results suggest that the weighted Euclidean-based pose estimation model successfully achieves moderately high prediction accuracy, but the exponential-based model results in poor orientation estimation accuracy. 

\begin{figure}[htb]
   	\centering
   	\includegraphics[width=0.95\textwidth]{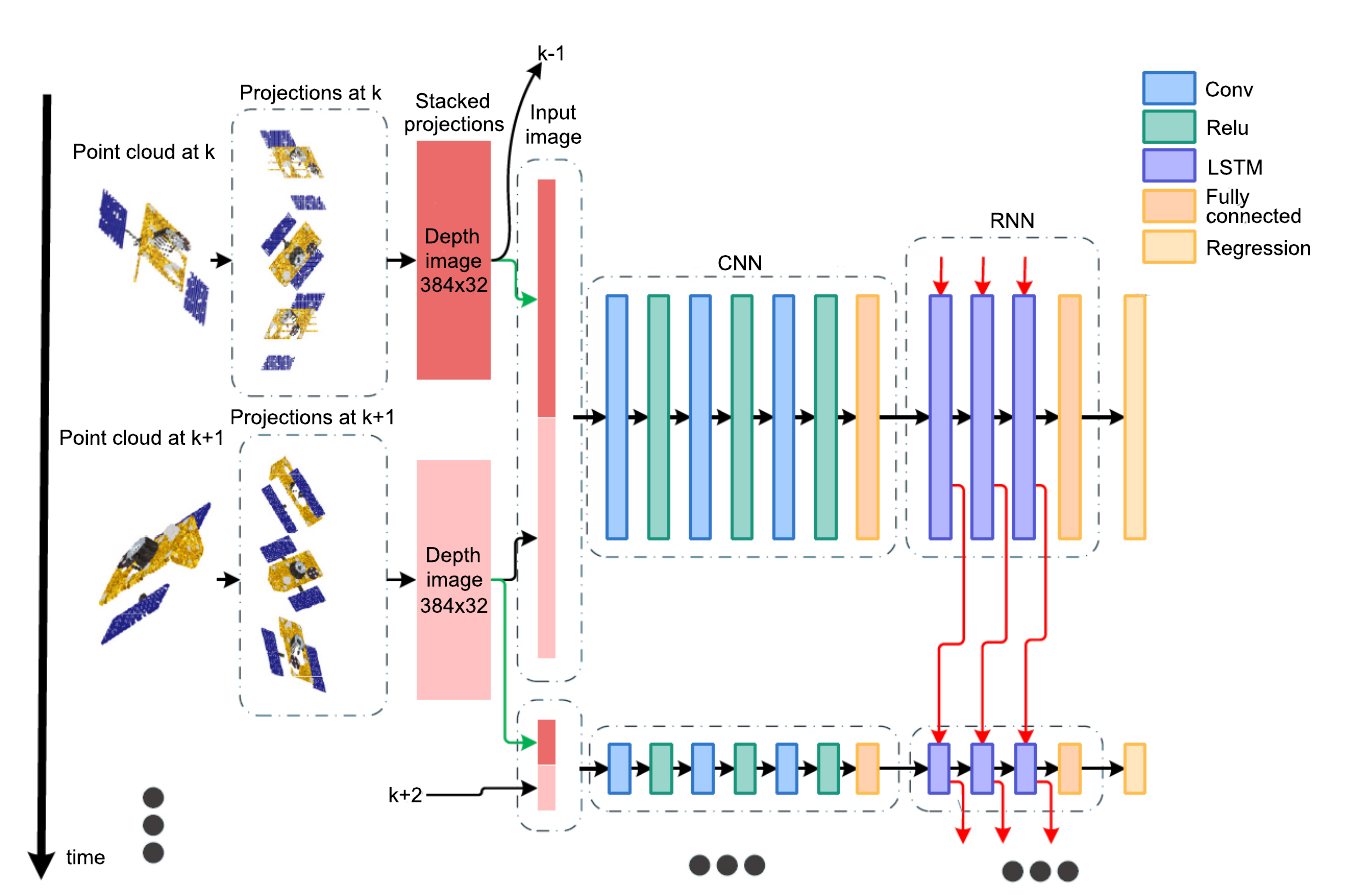}
 \caption{The \gls{drcnn} architecture of \citet{stamatis2020deeplo}. A shallow \gls{cnn} architecture is utilised to extract low-level features for projected images, which are then modelled with \glspl{lstm}.}
  \label{fig:supervised-stamatis2020deeplo}
\end{figure}

Instead of estimating poses at individual timesteps, \citet{stamatis2020deeplo} propose a \gls{drcnn} to regress the relative pose of spacecraft from frame to frame. For a relative spacecraft navigation system, these chained poses serve as continuous outputs, of which the continuity is vital to autonomous missions such as rendezvous and formation flyover. Specifically, the \gls{drcnn} consists of a \gls{cnn} module and followed a \gls{lstm} module to extract features of the input images and automatically modelling the relative dynamics, respectively (see \figref{fig:supervised-stamatis2020deeplo}). \gls{3d} lidar data is projected onto the image plane, yielding three different \gls{2d} depth images to be processed by a regular \gls{cnn}. As in \citref{wang2017deepvo}, the loss minimises the pose \gls{mse}, but the attitude is represented via a direction cosine matrix. Trials are conducted on both synthetic and real data. For the former, the Elite target satellite platform is used to create a self-occluded point cloud. The real dataset is acquired with a scaled mock-up of Envisat. Their results on both simulated and real lidar data scenarios demonstrate that the \gls{drcnn} achieves better odometry accuracy at lower computational requirements than current algorithms such as \gls{icp} \citep{besl1992method} and descriptor matching with $H_\infty$ filtering.

\citet{oestreich2020onorbit} study on-orbit relative pose initialisation by employing AlexNet-based transfer learning and a post-classification attitude refinement algorithm, which provides a foundation for future work in \acrconnect{cnn}{-based} spacecraft pose initialisation. Their research puts focus on answering several questions on the applicability of \gls{dl} to this domain, including the necessary amount of training imagery, attitude label discretisation, and the effects of lighting and image background on \gls{cnn} performance. Thus, AlexNet, used as the backbone of the proposed framework, only changes the final \gls{fl} to yield attitude labels. The output attitude, obtained from a single branch unlike \citref{sharma2019pose}, is then refined using the eight most likely labels via direction cosine matrix averaging. Synthetic images of the SpaceX Dragon capsule are rendered using Blender at a fixed range of \SI{20}{\metre}. Four different synthetic image sets are created to study the performance of the presented scheme and answer the proposed questions, namely considering a black and empty background, Sun angle variation, Earth background variation, and sensor noise. Based on their experimental results, it is indicated that: 
\begin{enumerate*}[label=\arabic*)]
\item both classification accuracy and attitude error exhibit an asymptotic trend;
\item the \gls{cnn} performs well in more challenging light conditions of the Sun variation dataset but poorly for the Earth background and sensor noise datasets; and
\item using the confidence rejection threshold in the refinement step can improve estimation accuracy slightly.
\end{enumerate*}

Recently, \citet{CosmasKenichi-21} first investigated the feasibility of \acrconnect{cnn}{-based} spacecraft pose estimation \textcolor{black}{by assessing the onboard inference capabilities of the model}. Accounting for power consumption and cost-effectiveness, the Xilinx Zynq Ultrascale+ multiprocessor \gls{soc} hybrid \gls{fpga} is proposed as a suitable solution. Two typical approaches and one presented framework are trained in Google Colab using the \gls{speed} dataset, showing that a U-Net-based detection network performs better than the ResNet-50 based direct regression scheme, albeit poorer than the developed ResNet34-U-Net model. Later, the ResNet34-U-Net pipeline is implemented on the proposed hardware, starting with a YOLOv3 for \gls{roi} detection, followed by a landmark localisation network to predict keypoints. Inference experiments, including an evaluation of the performance, compared to a desktop-based implementation, \gls{dl} processing unit resource utilisation, and power consumption are analysed with results of satisfactory accuracy and low on-chip power consumption of \SI{3.5}{\watt}.

To make a clear comparison between the aforementioned approaches, \Tabref{Table:summary-direct} summarises the surveyed \acrconnect{dl}{-based} direct frameworks for relative pose estimation. As shown, over half of the solutions employ transfer learning, which traditionally has also been considered by the most successful applications of \glspl{dnn} to terrain navigation problems. In terms of framework types, most are seen to adopt an estimation by regression or soft classifier. Open datasets of real images are limited; only PRISMA-25 and \gls{speed} are available. On the other hand, synthetic imagery can be simulated by different software or platforms, such as OpenGL, Gazebo, Unreal Engine 4, and Blender, leading to datasets such as \gls{urso}. Moreover, the recent research output volume demonstrates there is an increasing interest in \acrconnect{dl}{-based} spacecraft pose estimation, including even a first report on onboard implementations with \glspl{fpga}.

\begin{table}[t]
	\caption{Summary of \acrconnect{dl}{-based} direct pose estimation methods for spacecraft relative navigation. \label{Table:summary-direct}}
	\centering
	\resizebox{\textwidth}{!}{
	\begin{tabular}{llllll}
		\toprule 
		Ref.  &  Backbone  &  Transfer  & Type & Dataset & Comments\\
		&&learning&&&\\
		\midrule
		\citep{sharma2018pose} & AlexNet & ImageNet & Classifier & PRISMA, synthetic & Coarse initialisation \\
		\citep{sharma2019pose} & Faster \gls{rcnn} & ImageNet & Soft classifier & \gls{speed}, synthetic & Introduction of \gls{spn} \\
		\citep{sharma2020neural} & \gls{spn} & ImageNet & Soft classifier & \gls{speed}, PRISMA, synthetic (OpenGL) & Outperforms \citref{sharma2018pose} \\
		\citep{proenca2019deep} & ResNet-50 & \gls{coco} & Soft classifier & \gls{urso} (Soyuz \gls{sc}) & Soft classifier outperforms regressor in attitude\\
		\citep{hirano2018deep} & AlexNet & \ding{55} &  Regressor & Synthetic (Gazebo), real & Direct \gls{3d} keypoint regression\\
		\citep{arakawa2019attitude} & 2-layer \gls{cnn} & \ding{55} &  Regressor & Synthetic (Blender, JCSAT-3 \gls{sc}) & Evaluates robustness to noise, outputs quaternions\\
		\citep{sonawani2020assistive} & VGG-19 & ImageNet &  Regressor & Synthetic (Gazebo) & Cooperative object tracker \\
		\citep{PhisannupawongKamsing-22} & GoogLeNet & PoseNet &  Regressor & Synthetic (Unreal Engine 4, Soyuz \gls{sc}) & Comparison of two loss functions\\
		\citep{stamatis2020deeplo} & Shallow \gls{cnn} + \gls{lstm} & \ding{55} &  Regressor & Synthetic (Elite \gls{sc}), real (Envisat \gls{sc}) & Frame to frame motion estimator\\
		\citep{CosmasKenichi-21} & U-Net, ResNet, YOLOv3 & \ding{55} &  Regressor & \gls{speed} & Onboard \gls{fpga} implementation\\
		\citep{oestreich2020onorbit} & AlexNet & ImageNet &  Classifier & Synthetic (Dragon \gls{sc})  & Analysis of Sun angles, Earth presence, noise \\
		\bottomrule
	\end{tabular}}
\end{table}

\subsection{Indirect Frameworks for Spacecraft Relative Pose Estimation}
\label{sec-sub:indirectDLforPOSE}

Estimating the pose from images using end-to-end \acrconnect{dl}{-based} methods has been argued to yield inadequate feature representation and limited explainability, either of which has so far achieved subpar performances as opposed to geometry-based methods. \citet{Sattler2019} discuss the limitations of end-to-end \acrconnect{cnn}{-based} terrain pose regression and suggest that there is a gap for practical applications. \textcolor{black}{Moreover, the \gls{dnn} model has a risk of overfitting, which results in unpredictable drops in performance between the training images and test images due to memorising, rather than learning, properties of the former set that do not function well on the latter \citep{goodfellow2016deep}. Therefore, some research avenues have recently refocused on the indirect methods, which aim to combine \gls{dl} and conventional geometry-based techniques to refine the estimation of the pose.} 

\begin{figure}[t]
	\centering
	\begin{subfigure}[t]{0.65\textwidth}	\centering
		\includegraphics[width=\textwidth]{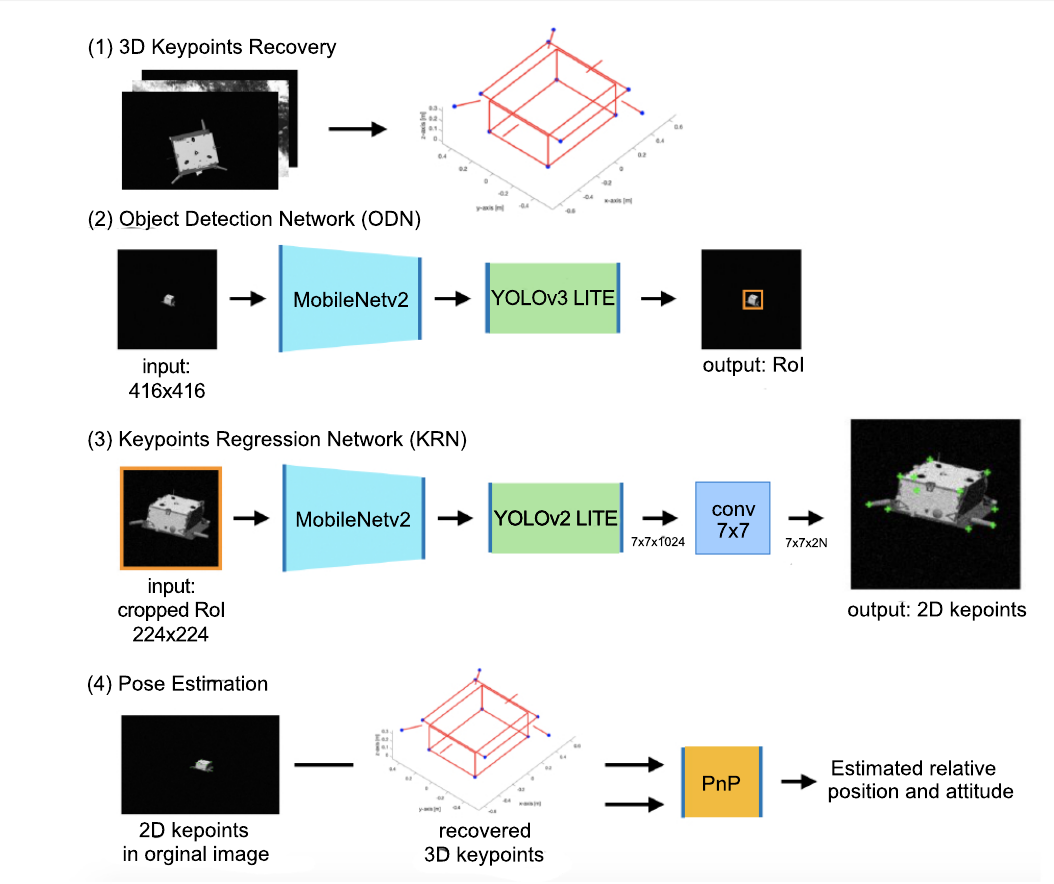}
		\caption{Pose estimation network}
		\label{fig:park2019a}
	\end{subfigure}\hfill
	\begin{subfigure}[t]{0.34\textwidth}	\centering
		\includegraphics[width = \textwidth]{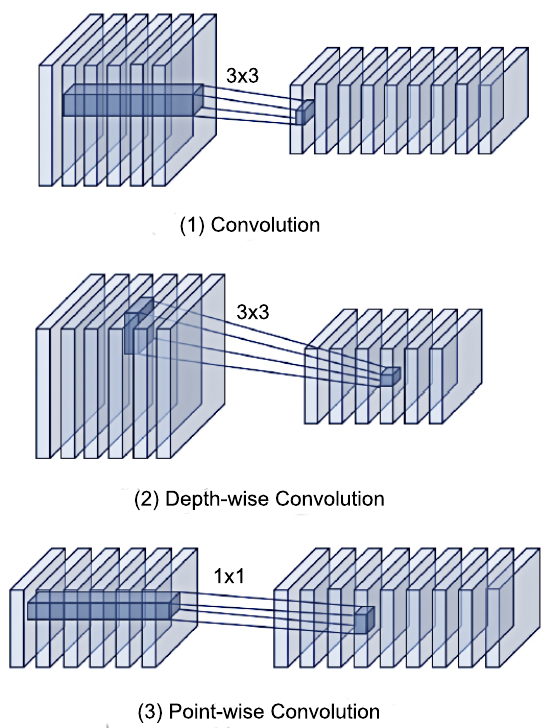}
		\caption{Different convolution operations}
		\label{fig:park2019b}
	\end{subfigure}
	\caption{Proposed \gls{dnn} framework in \citref{park2019towards} and comparison of three convolution operations.}
	\label{fig:park2019}
\end{figure}

To promote the practical use of \acrconnect{dl}{-based} pose estimation in space missions, \citet{park2019towards} take the \gls{spn} framework \cite{sharma2019pose} and modify it by employing both a novel \gls{cnn} for target detection and \gls{ransac} algorithms for solving the \gls{pnp} problem. The proposed \gls{cnn} is decoupled into the detection and pose estimation networks to determine the \gls{2d} bounding box of the \gls{roi} and to regress the \gls{2d} locataion of keypoints, respectively. As demonstrated in \figref{fig:park2019a}, the \gls{odn} and the \gls{krn} closely follow the pipeline of YOLOv2/YOLOv3 \citep{redmon2018yolov3}, but use MobileNetv2 \citep{sandler2018mobilenetv2} and MobileNet \citep{howard2017mobilenets}, respectively. To drastically reduce the number of network parameters, traditional convolution operations of the network are replaced with depth-wise convolutions followed by point-wise convolutions (see \figref{fig:park2019b}). 

Considering the lack of real space-based datasets with representative texture and surface illumination properties, \citet{park2019towards} also contribute with a new training procedure to improve the robustness of \glspl{cnn} to spaceborne imagery when trained solely on synthetic data. Inspired by \citref{geirhos2018imagenet}, they generate a new dataset by applying neural style transfer techniques \citep{huang2017arbitrary} to a custom synthetic dataset with the same pose distribution as \gls{speed}. After training with the new texture-randomised dataset, the proposed network performs better on spaceborne images and scores 4th place in \gls{kpec}.

The 1st place \gls{kpec} solution is also an indirect \acrconnect{dl}{-based} scheme proposed by \citet{chen2019satellite}, where \gls{dl} and geometric optimisation are combined to present a \acrconnect{cnn}{-based} pipeline for pose estimation from a single image. Firstly, \gls{3d} landmarks of the satellite are computed from the training set via multiview triangulation. A \gls{hrnet} \cite{sun2019deep} is then trained to regress the location of projected \gls{2d} corner point landmarks on the spacecraft from the input greyscale image. Finally, the optimal poses are obtained by the proposed geometric optimisation algorithm based on simulated annealing, where the initial pose is estimated from a \gls{pnp} solver. More specifically, the proposed \gls{dnn} framework contains two modules. The first uses an \gls{hrnet} front-end/Faster \gls{rcnn} combination to detect the \gls{2d} bounding box of the target in the input image. The \gls{roi} is then cropped and resized for use in the second model, which consists of a pure \gls{hrnet} and is trained on an \gls{mse} loss between the predicted and ground truth heatmaps of the visible landmarks in each image. 

\begin{figure}[htb]
     \centering
 	\includegraphics[width=0.85\textwidth]{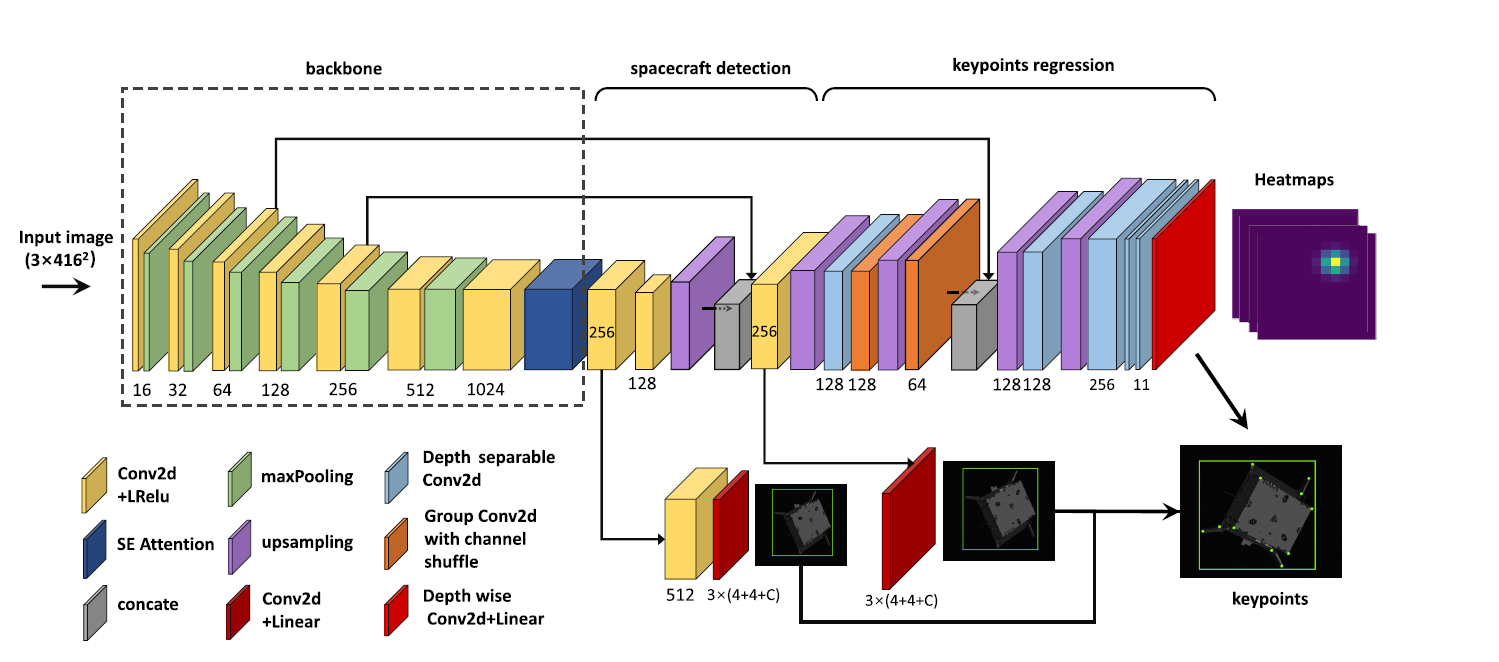}
 	\caption{The overall structure of the network designed by \citet{HuoLi-7}.}
 	\label{fig:huo2020}
\end{figure}

To achieve a fast and accurate estimate of the pose, \citet{HuoLi-7} developed a novel \glspl{dl}-based approach combining \gls{pnp} and geometric optimisation. A new and lightweight tiny-YOLOv3 based framework is designed to predict the \gls{2d} locations of the projected keypoints of the constructed \gls{3d} model. \figref{fig:huo2020} shows the corresponding regression network, in which the output of tiny-YOLOv3 is modified to establish a box reliability judgement mode for detecting the \gls{sc} and predicting the \gls{2d} \gls{roi}. Next, the regression of \gls{sc} keypoints is achieved by replacing the \glspl{fl} with \glspl{cl} to yield heatmaps. Finally, \gls{pnp} and bundle adjustment are utilised to generate the initial pose and optimise it, respectively, which improves the accuracy and robustness of the proposed approach. Their method is evaluated on the \gls{speed} dataset and achieves competitive performance in spacecraft pose estimation with a lighter computational footprint.

Another indirect \glspl{dl}-based scheme combines a \acrconnect{cnn}{-based} feature detector with a \gls{pnp} solver and an \gls{ekf} to guarantee a robust pose estimation \citep{cassinis2020cnn}. The authors build an hourglass-shaped \gls{cnn} composed of a six-block encoder and a six-block decoder to estimate the heatmaps of \num{16} predefined corners on the Envisat spacecraft. A target detection module is not incorporated since the presence of Earth in the background is not considered. Using the weights from the heatmaps, an associated landmark covariance is calculated. Two testing campaigns are then performed. The first one uses a dataset composed of singular images and computes the relative pose by incorporating the covariance of the regressed landmarks into the \gls{pnp} procedure \citep{ferraz2014leveraging}. The second campaign considers a sequential dataset simulating a V-bar approach with Envisat at a fixed relative distance, where the target performs a roll rotation with respect to the \gls{lvlh} frame of reference. The relative pose is estimated by a tightly coupled \gls{ekf} based on a Clohessy-Wiltshire dynamical model. Sensor measurements input of the filter, the landmark locations and covariances, come from the \gls{cnn}. The filter achieves steady-state position errors inferior to \SI{0.2}{\metre} for all axes, and the attitude errors are under \SI{2}{\degree}.

A pipeline similar to \citref{HuoLi-7} is investigated by \citet{HuanLiu-11}, achieving results nearly an order of magnitude better in the precision and accuracy of position and attitude estimation relative to the \gls{spn} framework. The training methodology consists of four steps:
\begin{enumerate*}[label=\arabic*)]
\item manual selection of images from the training dataset to be used for the reconstruction of the target \gls{3d} model,
\item detection of the \gls{2d} bounding box by an \gls{odn},
\item estimation of the \gls{2d} image location of keypoints from a \gls{krn}, and
\item projection of the \gls{3d} groundtruth keypoints onto the image plane and solving the \gls{pnp} problem from the correspondences with the estimated keypoints.
\end{enumerate*}
Differing from \citref{HuoLi-7}, the proposed target detection network and \gls{krn} in \citref{HuanLiu-11} apply the state-of-the-art \gls{hrnet} as backbone. The 6-\gls{dof} pose is finally predicted by non-linear minimisation of a Huber reprojection loss. The training dataset is constructed of synthesised greyscale images, and the test set images are captured in real-time using a monocular camera.

\citet{shi2018cubesat} transfer the state-of-the-art \gls{cnn} techniques to target CubeSat detection, but with no further discussion on pose estimation. Inception-ResNet-V2 \citep{szegedy2017inception} and ResNet-101 \citep{he2016deep} are combined and trained to estimate the bounding box of the target \gls{sc} with a laboratory test platform. The final \gls{fl} of ResNet-101 is reduced to two classes to differentiate between the \texttt{"1U\_CubeSat"} and \texttt{"3U\_CubeSat"} labels. The pre-trained weights of the \gls{cnn} are obtained from the \gls{coco} dataset \citep{lin2014microsoft} and further trained on a mixture of real and synthetic CubeSat images. Their simulation results indicate that the Inception-ResNet-V2 framework achieves a slightly higher accuracy and precision for \gls{sc} detection, whereas the ResNet-101 network is less computationally heavy. To tackle a similar problem, \citet{YangM2018} constructs a feature extraction network and \gls{rpn} structure framework based on Faster \gls{rcnn} \citep{ren2017faster} on the \gls{cnn} Caffe \citep{jia2014caffe} open platform. The proposed network performs intelligent identification of a spacecraft module in the image sequence and filters out the certain components of interest.

For other space missions beyond rendezvous, \citet{YiLin2020} studies the relative position estimation problem of a docking mission (below \SI{10}{\meter}). Assuming relative attitude has been adjusted, the method utilises a modified VGG-16 to regress the relative position between docking rings. Further position smoothing and relative speed estimation are achieved by Kalman filtering. 
Additionally, a satellite positioning error compensation technique based on \gls{dl} is discussed by \citet{ChenJM2018}. Large amounts of data are collected to generate a robust model; a \gls{cnn}, a depth belief network, and a \gls{rnn} are trained on satellite location data are collected by the Institute of Technology of the Chinese Academy of Sciences, which aims to generate a robust model. A \gls{cnn}, a depth belief network, and a \gls{rnn} are trained on the collected data, of which the \gls{cnn} performs the best compensation result.

\begin{table}[tb!]
	\caption{Summary of \acrconnect{dl}{-based} indirect pose estimation methods for spacecraft relative navigation. \label{Table:summary-indirect}}
	\centering
	\begin{threeparttable}[b]
	\resizebox{\textwidth}{!}{
	\begin{tabular}{llllll}
		\toprule 
		Ref.  &  Backbone  &  Transfer  & Type & Dataset & Pose estimation by\\
		&&learning&&&\\
		\midrule
		\citep{park2019towards} &YOLO, MobileNet& \ding{55} & Keypoint regressor & Synthetic (\glsxtrshort{nst}\tnote{1} ), \gls{speed} & \gls{pnp} \\
		\citep{chen2019satellite} & \gls{hrnet} + Faster \gls{rcnn} & \ding{55} & Keypoint regressor & \gls{speed} & \gls{pnp} \\
		\citep{HuoLi-7} & tiny-YOLOv3 & \gls{coco} & Keypoint regressor & \gls{speed}&\gls{pnp}\\
		\citep{cassinis2020cnn} & Hourglass network & \ding{55} &  Keypoint regressor & Synthetic (Cinema 4D, Envisat \gls{sc}) & \gls{pnp} + \gls{ekf} \\
		\citep{HuanLiu-11} & 2-layer \gls{cnn} & \ding{55} &  Keypoint regressor & Synthetic, real (lab) & \gls{pnp}\\
		\citep{YiLin2020} & VGG-16 & \ding{55}&  Position regressor & Synthetic (Blender) & \gls{ekf}\\
		\citep{shi2018cubesat} & ResNet & \gls{coco} & Classifier & Synthetic, real & CubeSat detection \\
		\citep{YangM2018} & Faster \gls{rcnn} &\ding{55} & Regressor  & Synthetic & Object detection\\
		\citep{ChenJM2018} & \gls{lstm} &\ding{55} & Regressor  & Real &  Position error compensation  \\
		\bottomrule
	\end{tabular}}
	\begin{tablenotes}    
        \footnotesize               
        \item[1] \glsxtrlong*{nst}; applied to randomise the texture of the spacecraft. 
        \end{tablenotes}          
	\end{threeparttable}  
\end{table}

\Tabref{Table:summary-indirect} contains a brief summary of
indirect \acrconnect{dl}{-based}\ algorithms for spacecraft pose estimation and related applications. As illustrated, the earlier studies (last four referenced) are more focused on parts of pose estimation missions, such as detection and position estimation. \gls{pnp} and \gls{ekf} are commonly combined to refine poses output by \glspl{dnn}. Moreover, transfer learning and very deep networks are rarely utilised when \gls{dl} methods are combined with optimisers. This could potentially be due to the fact that the pipelines rely heavily on these optimisation steps at the end, which are able to guarantee a decent estimate of the pose. In this way, shallow networks can reduce the computational cost, which is beneficial for practical use and potential onboard implementations.
\section{Crater and Hazard Detection for Terrain Navigation Using \glsfmtshort{dl}}
\label{sec:craterdetection}
\begin{figure}[t]
    \centering
	\includegraphics[width=0.7\textwidth]{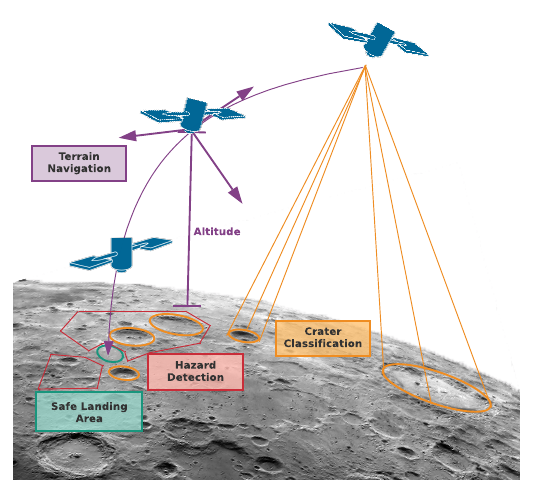}
	\caption{Scenarios of crater detection, hazard detection for safe landing area, and terrain navigation.}
	\label{fig:diagramsFORcratersAPP}
\end{figure}

\noindent Exploring and landing on the lunar surface has long been a challenge of great interest within space technology and science. Recent developments in \gls{dl} have led to a renewed interest in learning-based \gls{trn}. Craters are ideal landmarks for relative navigation on or around the Moon and asteroids \citep{wang2018crateridnet,vaniman1991lunar}. Additionally, hazards should be avoided for a successful landing mission. This section, therefore, reviews the field of \acrconnect{dl}{-based} terrain navigation in three aspects: crater detection, hazard detection, and \gls{trn} methods, all using \glspl{dnn}. 
\textcolor{black}{\Figref{fig:diagramsFORcratersAPP} shows schematic diagram and difference between the three aspects.}

\subsection{Crater detection}
\noindent With advances in computer vision and successful applications of \glspl{cnn} in the object detection area, CNN-based crater detection algorithms are also emerging. However, most of the earlier methods only utilise a \gls{cnn} as a classifier to validate selected features, such as in \citrefpl{emami2015automatic,cohen2016crater,palafox2017automated}.
The shapes of natural craters vary in morphology, including peak rings, central pits, and wall terraces \citep{Keefe1999Complex}. Some craters may also overlap with others. Considering the illumination conditions and different poses of on-board cameras, the imaged craters can be diverse in terms of dimensions and appearance \citep{wang2018crateridnet}. 
Conversely, robust crater detection algorithms have been developed by applying \glspl{dnn} to fully process raw crater images, exhibiting promising results which have attracted a lot of interest.

The \gls{pycda} \citep{klear2018pycda} is an open-source crater detection library composed of a detector, extractor, and classifier, which focuses on detecting new craters that have never been catalogued. \gls{pycda} uses a downsized U-Net architecture to compute the per-pixel likelihoods of a crater rim from inputs of greyscale intensity images. The pixel prediction map is then fed to the extractor to generate a list of crater candidates. A classifier \gls{cnn} is finally applied to determine true craters. Thanks to \gls{pycda}, a considerable amount of craters have been detected and categorised, thus helping to generate new labelled datasets for training and testing of \gls{dl} algorithms.

\begin{figure}[t]
    \centering
	\includegraphics[width=0.9\textwidth]{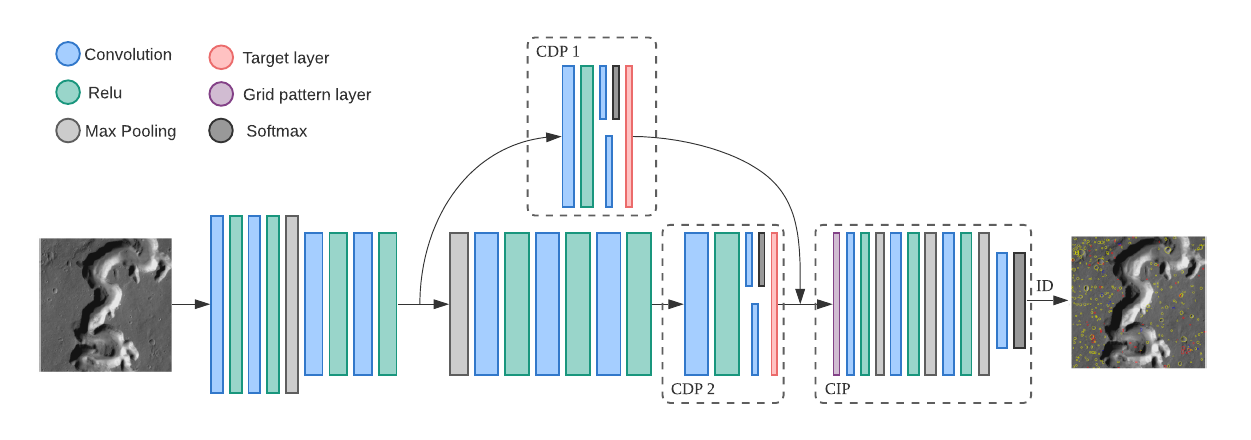}
	\caption{Framework of CraterIDNet reproduced from \citet{wang2018crateridnet}.}
	\label{fig:wang2018}
\end{figure}

\citet{wang2018crateridnet} proposed an end-to-end fully \gls{cnn}, CrateIDNet, for simultaneous crater detection and identification. CraterIDNet takes remote sensing images in various sizes and outputs detected crater positions, apparent diameters, and indices of the identified craters. Instead of using large off-the-shelf \gls{dnn} models, a small \gls{cnn} architecture pre-trained on Martian crater samples \citep{robbins2012new} is first developed to extract feature maps. Next, two pipelines, namely \gls{cd} and \gls{ci} are proposed for simultaneous detecting and identifying craters. 
\textcolor{black}{The \gls{cd} process involves detecting the presence of craters and locating them within the image if they exist. The output of \gls{cd} is then fed to the \gls{ci} process to match the detected craters to surface landmarks in a known database, and matches of \gls{ci} will  provide position estimation.}
\figref{fig:wang2018} shows the whole framework of CraterIDNet. 
The \gls{cd} modifies the \gls{rpn} architecture \citep{ren2017faster} as the backbone, regressing objectness scores and crater diameters from feature maps. Due to different craters sizes, two \gls{cd} pipelines are designed by sharing same \glspl{cl} but with different parameters. Later, craters are identified by \gls{ci} that combines a proposed grid pattern layer and \gls{cnn} framework. For the training and testing dataset, \num{1600} craters are manually catalogued and enlarged to a final sample set of \num{16000} instances through data augmentation. Experiments reveal that the light CraterIDNet with a size of \SI{4}{\mega\byte} performs better than previous algorithms \citep{cohen2016crater}.

\begin{figure}[t]
    \centering
	\includegraphics[width=0.9\textwidth]{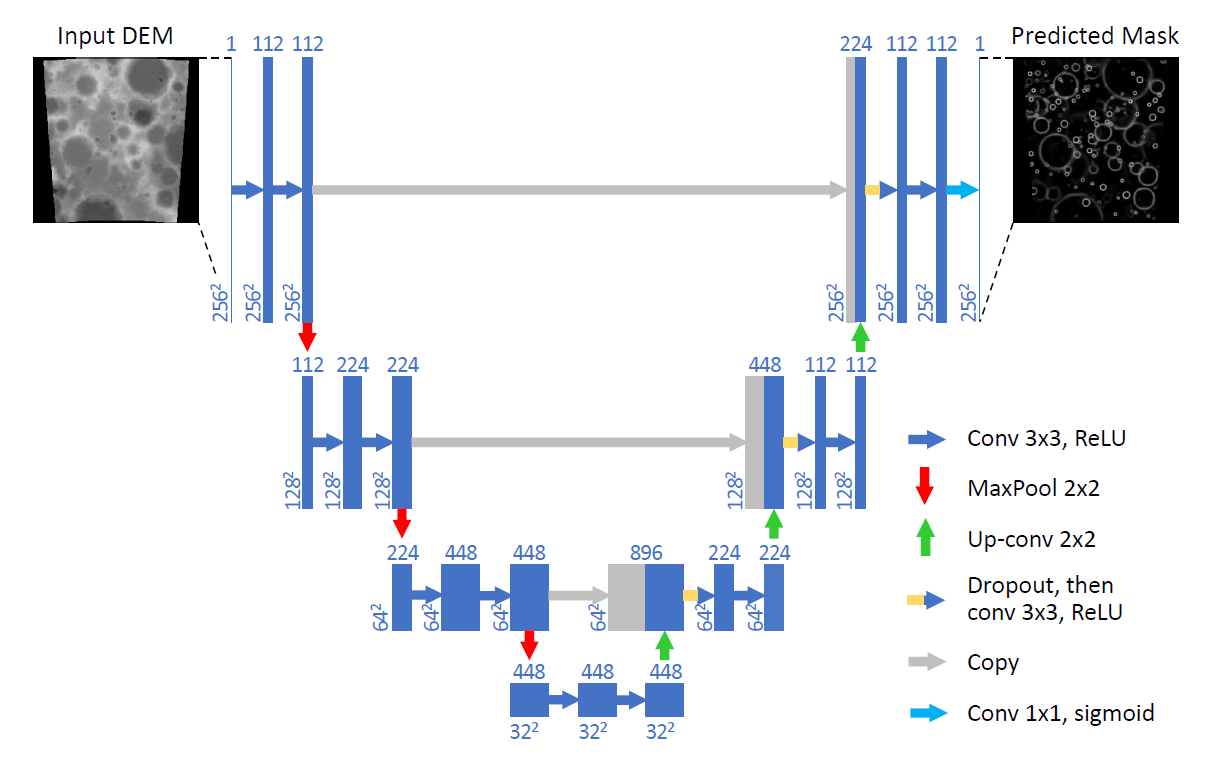}
	\caption{Architecture of DeepMoon network \citep{silburt2019lunar}}
	\label{fig:deepmoon2019}
\end{figure}

\citet{silburt2019lunar} employ a \gls{cnn} architecture for robust crater detection on the lunar surface using \glspl{dem}. The method relies on the developed DeepMoon network to identify the craters in terms of their centroid and radius, and outputs pixel-wise confidence maps of crater rims on the surface of a rocky body. DeepMoon modifies U-Net \citep{ronneberger2015unet} by changing the input image size, the number of filters in each convolution layer, and the use of dropout \citep{srivastava2014dropout} in the expansive path for memory limitations and regularisation respectively. \figref{fig:deepmoon2019} presents the DeepMoon architecture.

For training, the data used in DeepMoon is generated by merging two human-generated crater catalogues, which are the \gls{lro} \gls{wac} Global Lunar \gls{dem} \citep{povilaitis2018crater} and the \gls{lro} Lunar Orbiter Laser Altimeter \gls{dem} \citep{head2010global}. The dataset is split into equal train-validation-test parts, yielding \num{30000} \gls{dem} images per part. The minimised loss function is chosen as the pixel-wise \gls{bce}. 
DeepMoon produces a crater rim prediction mask, which is then fed to a low-level image process and a template matching procedure to determine the actual craters. The median fractional longitude, latitude and radius errors are \SI{11}{\percent} or less, representing good agreement with the human-generated datasets. Additionally, transfer learning from training on lunar maps to testing on maps of Mercury is qualitatively demonstrated successfully.

\begin{figure}[t]
    \centering
	\includegraphics[width=0.9\textwidth]{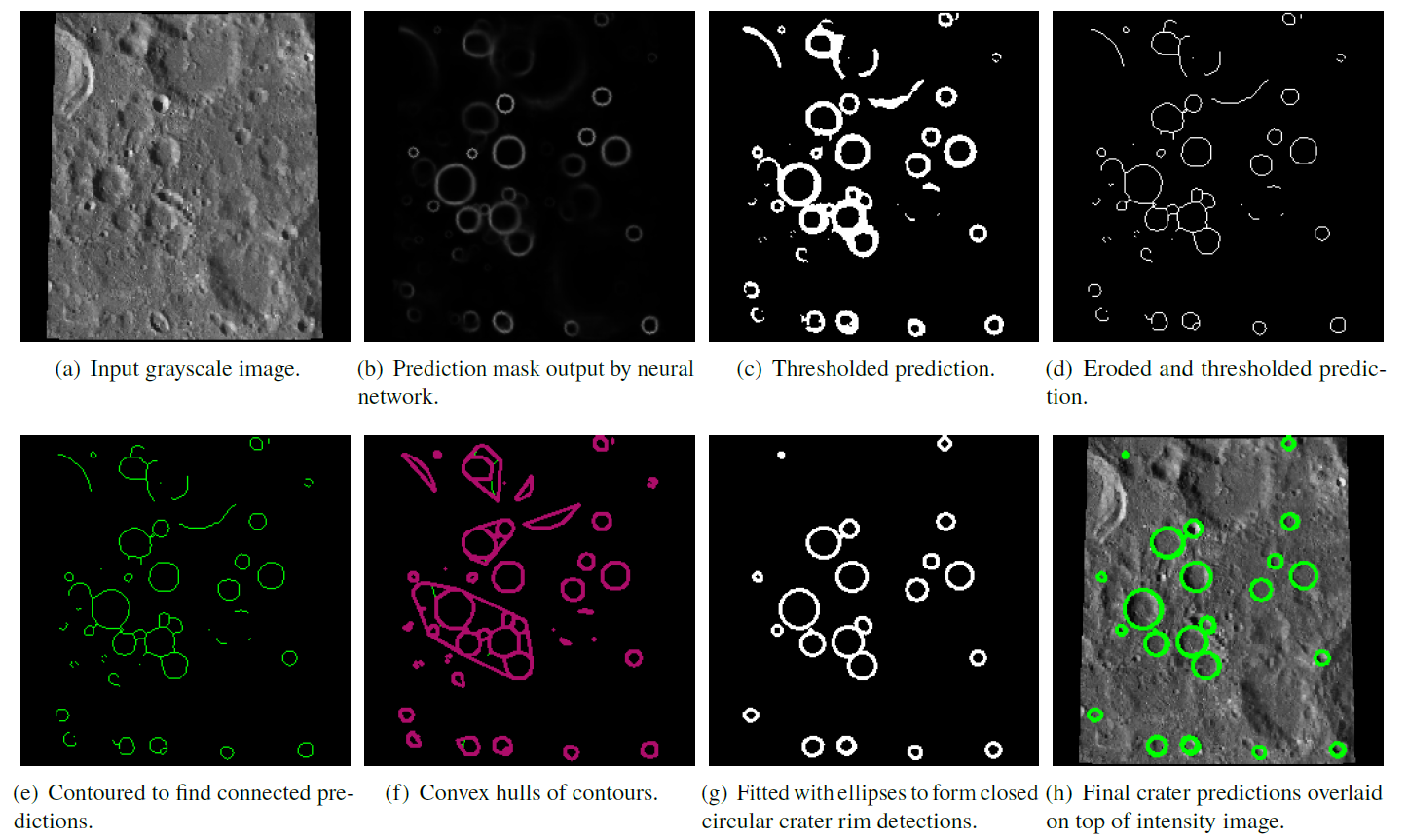}
	\caption{Feature extraction steps of LunaNet \citep{downes2020}}
	\label{fig:lunanet2020}
\end{figure}

\citet{downes2020} propose the LunaNet framework to detect craters for lunar \gls{trn}, which is quite similar to DeepMoon with the exception that it takes greyscale images as inputs. Thus, the method is more suitable for implementation aboard a spacecraft equipped with an optical camera without the need for a depth sensor. The output of the \gls{cnn} is, like DeepMoon, a crater rim prediction mask. However, the craters are extracted through a different method and, \figref{fig:lunanet2020} shows each feature extraction step of the LunaNet, including prediction mask, eroded and thresholded prediction, contour detection, and ellipse fitting. The data preparation is also akin to the process followed by DeepMoon, with the \gls{lro} \gls{wac} Global Lunar \gls{dem} dataset \citep{povilaitis2018crater}, followed by a histogram rescaling of the input greyscale images to match the intensity distribution of a \gls{dem} image. Based on the pre-trained DeepMoon weights, LunaNet reduces the training effort and final detection results. Experimental results indicate that LunaNet's performance surpasses DeepMoon and \gls{pycda} in terms of robustness to noisy images, location accuracy, and average crater detection time.

\begin{figure}[t]
    \centering
	\includegraphics[width=\textwidth]{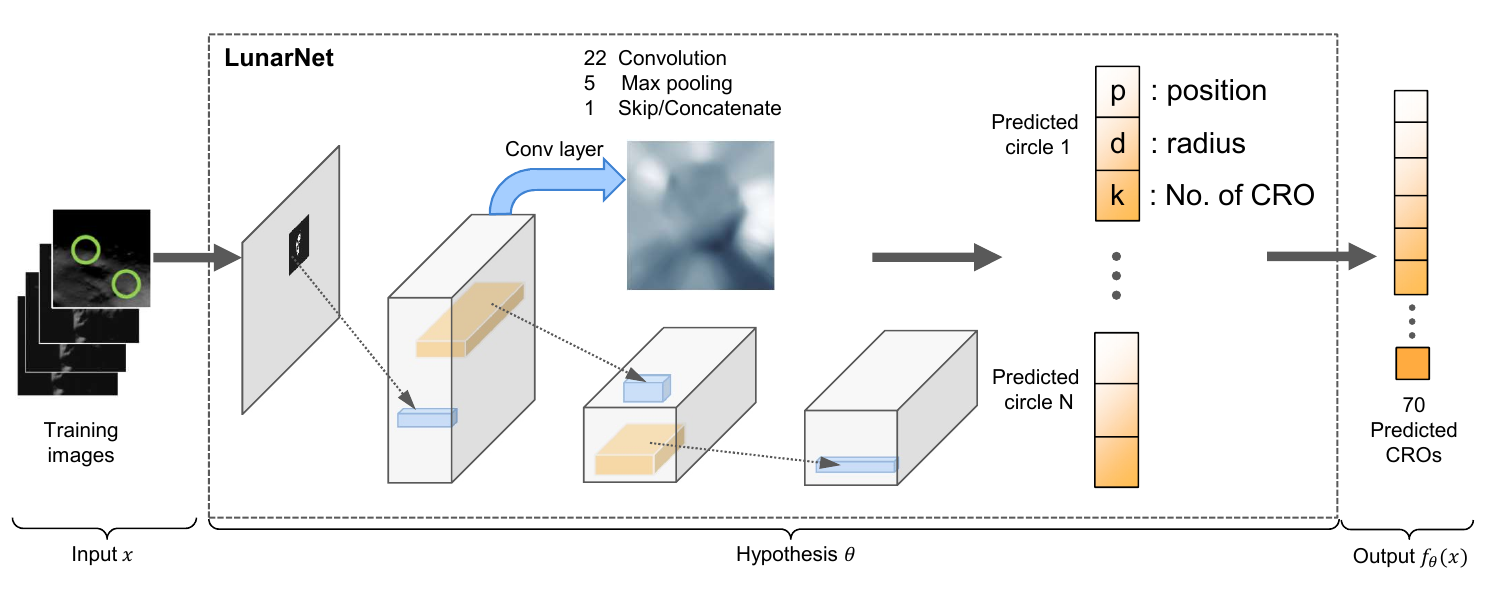}
	\caption{\acrconnect{cnn}{-based} \gls{cro} discriminator (LunarNet) \citep{Lee2020Deep}}
	\label{fig:lee2020}
\end{figure}

\textcolor{black}{It has been observed that areas with low solar angles, where there is heavy shadowing, result in reduced crater detection reliability.}
\citet{Lee2020Deep}
employ a \acrconnect{cnn}{-based} object detector to distinguish likely landmark candidates and predict detection probabilities along various lighting geometric flight paths, aiming to identify high-value landmarks by using optical navigation systems. A massive dataset based on real lunar-surface data is collected.
A \gls{cro} is defined as an image object with specific latitudes and longitudes. The LunarNet architecture (\figref{fig:lee2020}; see also the process of LunarNet-based landmark selection in \citref{Lee2020Deep}) is then used and trained to identify \glspl{cro} by maximising the discrimination between local areas of the Moon. 
Finally, the \gls{cro} performance map is formed based on the scored \glspl{cro} arranged by considering the azimuth and elevation angles of the Sun during the year. Numerical experimental results demonstrate that the proposed landmark detection pipeline can provide usable navigation information even at Sun angle elevations of less than \SI{1.8}{\degree} in highland areas, which indicates a successful application for the worst dark highlands near the South Pole.

\subsection{Hazard detection}
\noindent Hazard detection is considered the vital research field of space \gls{trn}
to avoid failures during landing. In 1974, Apollo program officials introduced manual hazard and target selection for lunar descent guidance \citep{klumpp1974apollo}. In the past decade, many algorithms have been developed which benefit from the increasing computational power of processor devices. Since 2006, promoted by the \gls{alhat} project \citep{epp2008autonomous} conducted by the \gls{nasa}, there has been growing interest in hazard estimation based on \glspl{dem}.
In 2012, \citet{furfaro2012autonomous}  implemented an \gls{ai} system to autonomously select a soft landing site in a region with safe terrains for Venus and Titan. 
In 2015, \citet{maturana20153d} used what they called a \gls{3d}-\gls{cnn} to create a safety map for autonomous landing zone detection for helicopters.

Earlier research in \acrconnect{nn}{-based} \gls{hda} for lunar landing is studied by  \citet{lunghi2014autonomous} and \citet{lunghi2016multilayer}, who demonstrate the ability and attractive properties of \glspl{ann} for real-time applications.
The ground truth is calculated from the corresponding \gls{dem} by thresholding pixel-wise figures.
Input images of the terrain are manually processed at a resolution of \SI{1024x1024}{px} to extract a \num{13}-dimensional vector per pixel comprising the image intensity mean, standard deviation, gradient and the \gls{log} at three different scales, and the Sun's inclination angle. Following this, the crafted features are fed to a neural network, outputting a \SI{256x256}{px} hazard map with each pixel value denoting a confidence value. From the output hazard map, candidate landing sites are obtained via pixel thresholding and scored global landing potential by analysing minimum radial dimension requirements, distance to an a priori nominal landing site, and the \gls{nn} scores of pixels inside the candidate radius. The target landing site is selected as the one that maximises the global score. 
Two different pipelines are developed: one based on a \gls{mlp} with \num{15} nodes, and the other based on a cascading \gls{nn} with successive layers of hidden information added during training. A test set of \num{8} images including four landscapes in two Sun inclination angles are utilised to evaluate two proposed pipelines. The predicted hazard maps during training have a negligible difference, with \num{0.0194} for the \gls{mlp} and \num{0.02039} for the cascade of pixel-wise \gls{mse}. However, the former proved better at determining safe landing sites. In addition, qualitative results have been presented for asteroid images acquired by the Rosetta probe.

\begin{figure}[t]
    \centering
	\includegraphics[width=0.85\textwidth]{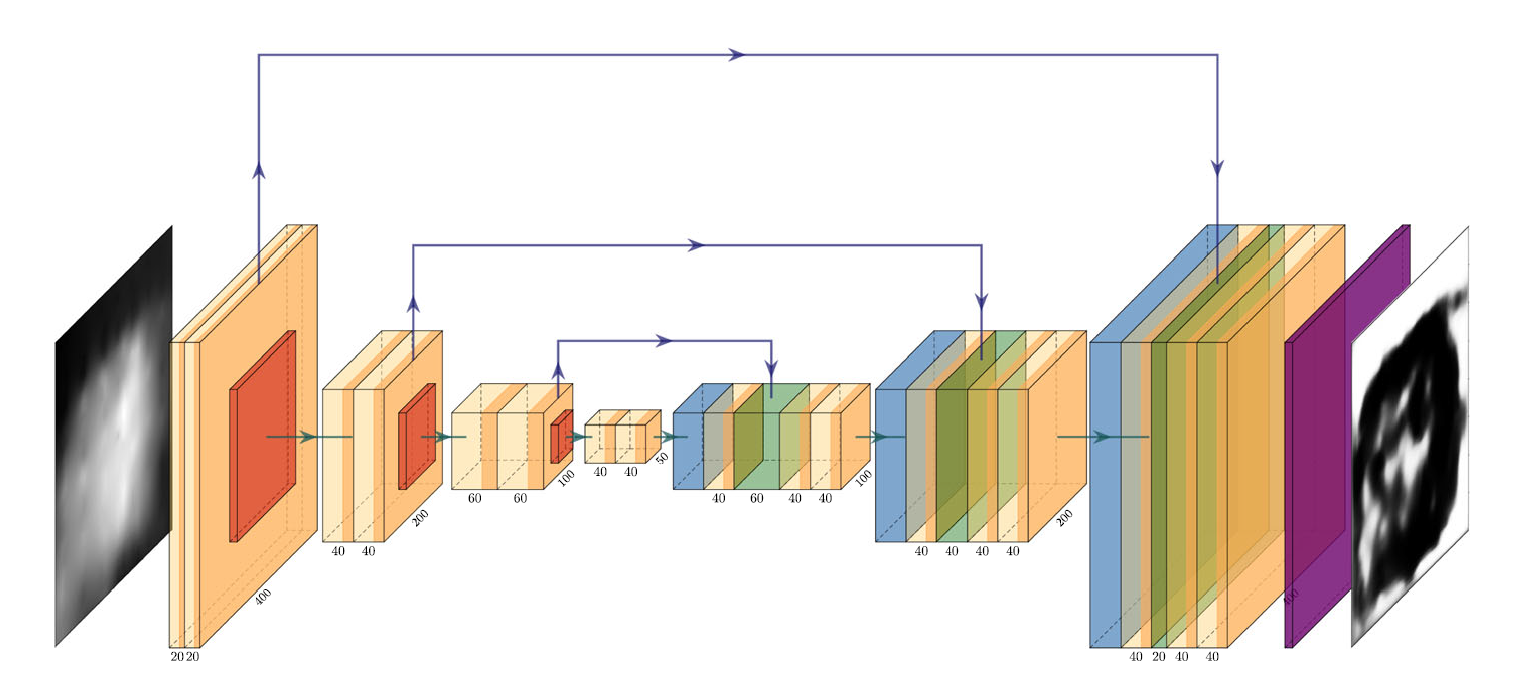}
	\caption{The topology of the network in \citep{moghe2020deep}}
	\label{fig:moghe2020}
\end{figure}

Recently, \citet{moghe2020online} presented a more modern approach towards tackling the same problem. Aiming at the hazard detection of the \gls{alhat} project, the authors implement an hourglass-like \gls{cnn} architecture with copy and crop connections based on U-Net  \cite{ronneberger2015unet}.
The framework processes \glspl{dem} directly and classifies safe and hazardous landing spots with the output map. Through data augmentation and transforming existing datasets, they create a new dataset from the \gls{lro} dataset \citep{emami2015automatic}.
The output, similarly to \citref{lunghi2016multilayer}, is a confidence map followed by a threshold to yield a binary landing/non-landing score, despite not provide a specific landing site. Results on a set of \num{100} testing images demonstrate an average hazard mapping Dice accuracy score of \SI{83}{\percent} and indicate the potential of real-time processing in future missions. 
Later, \citet{moghe2020online} expand and modify their work in \citref{moghe2020deep}, using the same network architecture but featuring improved layers covering the input size, output size, and layer width. The topology of the modified network is illustrated in \figref{fig:moghe2020}. Similarly, the Albumentations data augmentation suite \citep{buslaev2020albumentations} is used to prepare data. The modified \acrconnect{cnn}{-based} network outputs a mean pixel accuracy of \SI{\sim 92}{\percent} on the same testing dataset of \citref{moghe2020online}.

\subsection{Terrain navigation}

\noindent In the field of image-based planetary \gls{trn}, \citet{campbell2017deep} first utilise a \gls{cnn} architecture trained on a series of images rendered from a \gls{dem} simulating the Apollo 16 landing site to output the position of a spacecraft relative to the ground along one direction. The problem is posed by taking a centre \SI{128}{px} wide strip from the original \SI{1024x1024}{px} nadir base image, considering each pixel location along the on-track dimension as its own class. \SI{128x128}{px} training images are generated by sampling every \SI{8}{px} horizontally across the strip and rendering it \num{11} times at different Sun illumination angles. The \num{1024}-dimensional one-hot vector, which labels the position along the track line, is then applied to each image. The \gls{cnn} is composed of three \glspl{cl} and each followed by a max pooling layer. Thirty images are rendered at unseen Sun angles to make up a test dataset. Six of these are classified correctly, while in general, the maximum error observed is equal to \SI{5}{px}. For a ground sample distance of \SI{0.5}{\metre}, this means that achieved position errors are bounded at \SI{2.5}{\metre}. The testing is repeated for training images resampled at \SI{4}{px}, and the errors dropped to a maximum of \SI{3}{px} (or \SI{1.5}{\metre}).

In 2020, \citet{Lena8} explored how their LunaNet could be applied to the \gls{trn} problem and reported a system for the robust estimation of relative position and velocity information. Thus, LunaNet is utilised to detect and match craters to known lunar landmarks from frame to frame across a trajectory. The matched craters are treated as features feeding to a feature-based \gls{ekf}, where the state of the filter is the position and velocity of the camera in \gls{lclf}, as well as the location of detected features in this same reference frame. Compared to an image processing-based crater detection method \citep{singh2008lunar}, the LunaNet + \gls{ekf} combination produces considerable improvements on the accuracy of the \gls{trn}, with reliable performance in variable lighting conditions.

\begin{figure}[bt]
\centering
  	\begin{subfigure}[t]{0.39\textwidth}	\centering
		\includegraphics[width=0.99\textwidth]{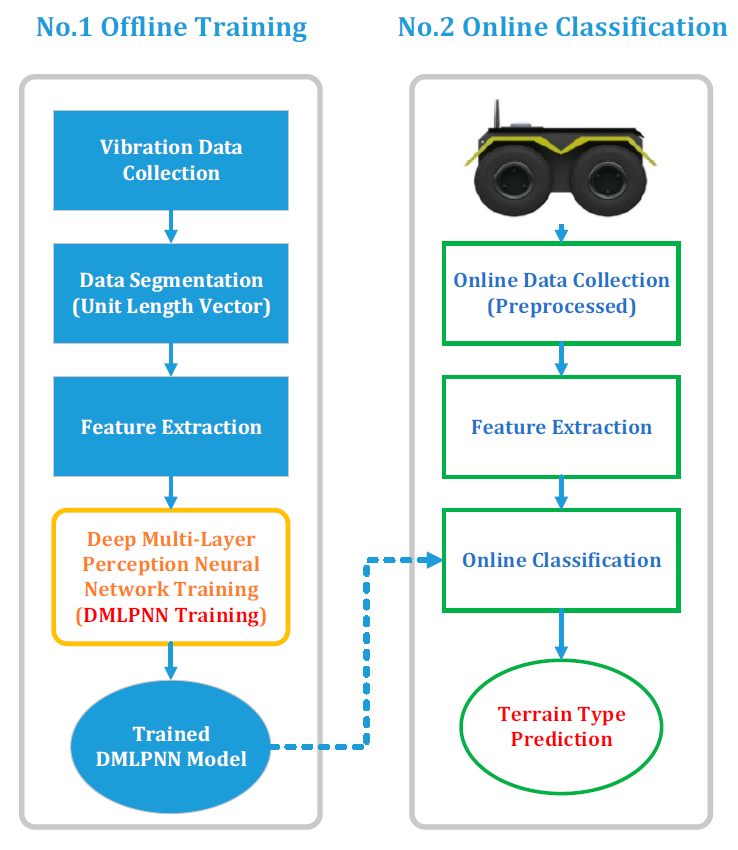}
		\caption{\acrconnect{nn}{-based} classification flow}
		\label{fig:bai2019a}
	\end{subfigure}
  	\begin{subfigure}[t]{0.59\textwidth}	\centering
		\includegraphics[width=0.99\textwidth]{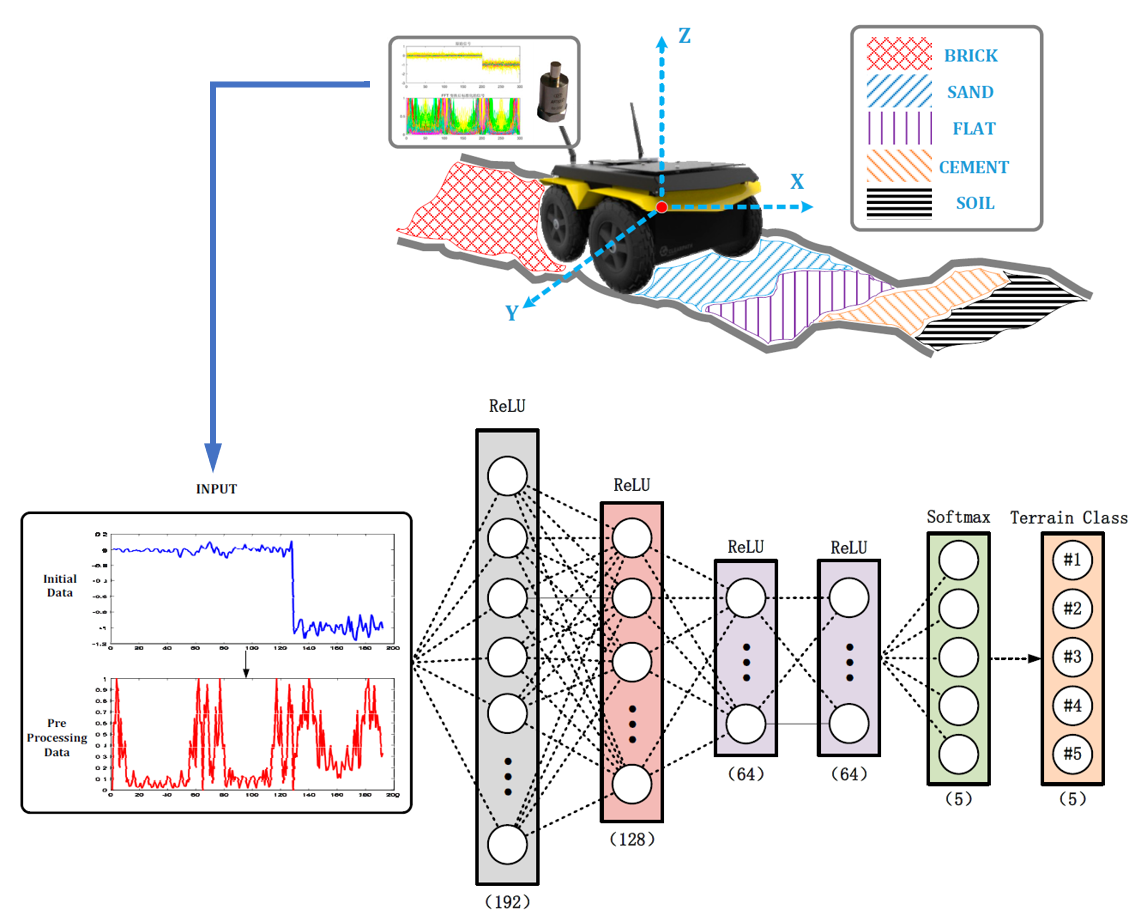}
		\caption{The structure of \acrconnect{dmlp}{\gls{nn}}}
		\label{fig:bau2019c}
	\end{subfigure}
	\caption{\acrconnect{dmlp}{\gls{nn}} classification flow and proposed architecture of \citet{bai2019deep}.}
	\label{fig:bai2019}
\end{figure}

Accurately identifying the detected terrain environment helps to achieve successful missions relying on planetary rovers. However, vision-based \gls{trn} systems are difficult to effectively perceive the material and mechanical characteristics of the terrain environment. Thus, \citet{bai2019deep} and \citet{BaiCC2019} investigate several terrain classification and recognition methods from vibration using \glspl{dnn}. The experimental and \gls{nn} classification flows are illustrated in \figref{fig:bai2019a}. The authors compare three different learning-based approaches towards terrain material perception and classification: an improved \gls{nn} algorithm, a \acrconnect{dmlp}{\gls{nn}} algorithm, and \gls{cnn}-\gls{lstm} based algorithm. Among these three schemes, the \acrconnect{dmlp}{\gls{nn}} achieves the best performance \citep{bai2019deep}. To classify textures, \acrconnect{dmlp}{\gls{nn}} (shown in \figref{fig:bau2019c}) adopts a five-\gls{fl} architecture, in which the activation functions are ReLU and softmax for the first four and last layers, respectively. For the dataset and training, three-dimensional raw vibration data collected by sensors is first segmented to a vector with a fixed duration. Using the fast Fourier transform, the vector is then transferred to the frequency domain, in which the eigenvectors are obtained for network training. Five different textures, including brick, sand, flat, cement, soil, are trained and recognised by \acrconnect{dmlp}{\gls{nn}} with high overall classification accuracy.

\begin{figure}[tb]
    \centering
	\includegraphics[width=0.95\textwidth]{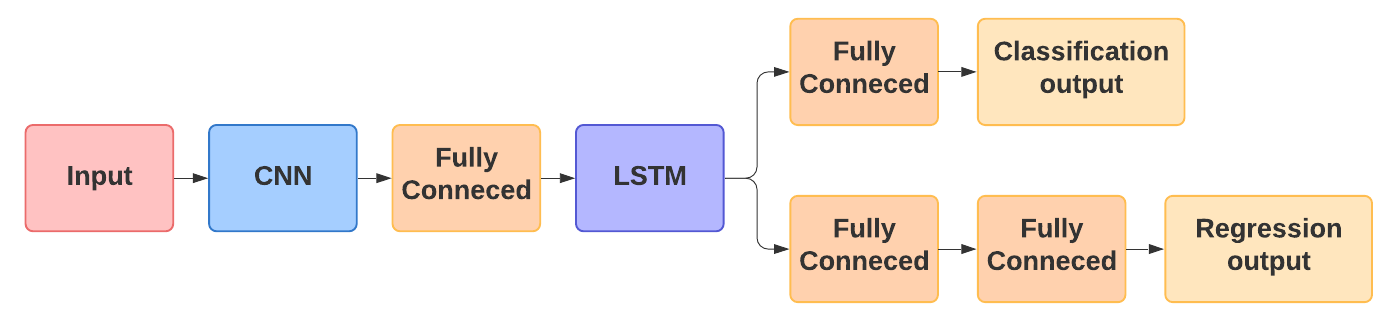}
	\caption{The architecture proposed by \citet{furfaro2018deep}}
	\label{fig:furfaro2018deep}
\end{figure}

For an autonomous lunar landing scenario, \citet{furfaro2018deep} propose a \gls{dnn} architecture that predicts the fuel-optimal control actions only using raw greyscale images taken by an on-board lander camera (\figref{fig:furfaro2018deep}). The architecture is a five-layer \gls{cnn} with three sequential images as input for each timestep. The \gls{dnn} is modified with an \gls{lstm} back-end connected to two further branches: one for regression and one for classification. For training the network, a set of optimal trajectories is computed numerically via Gauss
pseudo-spectral sampling methods using the \gls{gpops} \citep{patterson2014gpops}, producing a set of initial and final relative positions and velocities. Each state of the optimum trajectory is simulated by raytracing a \gls{dem} of a patch on the Lunar surface, resulting in \num{562} images with \SI{256}{px} $\times$ \SI{256}{px} of resolution. For better performance, the model is retrained explicitly with subsets of data that do not produce satisfactory results on the first try \citep{Furfaro2016Relative}.

\subsection{Brief summary}

\noindent The use of \glspl{dnn} in crater and hazard detection has not been widely investigated due to the lack of labelled databases. Datasets containing crater images which are open to the public do exist, e.g. the Lunar Crater Database\footnote{\url{https://astrogeology.usgs.gov/search/map/Moon/Research/Craters/lunar_crater_database_robbins_2018}.} \citep{emami2015automatic} or the Robbins Mars Crater Database \citep{robbins2012new}. Yet, manually catalogued craters are required for applying supervised \gls{dl} methods as presented in \citep{wang2018crateridnet,silburt2018deepmoon,downes2020,povilaitis2018crater,head2010global}. There exists still a gap towards the automatic generation of \gls{dl} crater datasets, and in the past two years there have been increasing studies of \glspl{dnn} with promising performance for \gls{trn} tasks.

\section{\glsfmtshort{dl}-based Relative \textcolor{black}{Navigation} for Asteroid Research}
\label{sec:asteroid}
\subsection{Challenges and Motivations for \glsxtrshort{dl}-based Asteroid Exploration}

\noindent Recent trends in small planetary exploration have led to a proliferation of studies that include asteroids and comets, pushed by scientific, planetary defence, and resource exploitation motivations \citep{pugliatti2021navigation,beauchamp2018technology}. Autonomous navigation is demanded due to the long communication delay and complicated dynamic environment in the vicinity of asteroids \citep{shuang2008landmark}.
Thus, it becomes necessary to develop new autonomous navigation algorithms for future asteroid sample and return missions, for which \gls{dl} techniques may provide a potential alternative.

The aforementioned studies demonstrate the potential of \glspl{dnn} for image patch classification invariant under illumination changes applied to terrain navigation. The same principle could be used for other relative navigation applications, such as asteroid location pinpointing, illustrated in \figref{fig:intro-scenarios-ap}. \gls{nea} missions, however, are more challenging than lunar missions; this is because one has limited information on the gravitation and environment of asteroids. If the celestial body and its orbit environment are in great uncertainty, all plans elaborated on-ground may dramatically fail when implemented in space \citep{guffantimulti2018multi}. 
Additionally, the lack of labelled ground truth data for asteroids challenges the application and development of \gls{dl} techniques in asteroid detection and landing \citep{ravani2021site}. 

\subsection{Previous Works Contributing to the Field}

\noindent For asteroid missions, earlier researchers have made various contributions towards \acrconnect{nn}{-based} orbit and dynamics uncertainty estimation. \citet{harl2013neural} develop a \acrconnect{nn}{-based} state observer to estimate gravitational uncertainties that spacecraft experience in an asteroid orbiting scenario. The \gls{nn} of the proposed state observer outputs the uncertainty as a function of the states instead of discrete values of an \gls{ekf}. 
\citet{guffantimulti2018multi} trains a neural network as an autonomous motion planning unit to compute the optimal spacecraft orbital configuration, which takes the uncertain \gls{nea} dynamics parameters created by navigation filters and the selected trade-off. 
\citet{song2019fast} also employ a six-hidden-layer \gls{dnn} to quickly estimate the gravity and gradient of irregular asteroids and further apply the \gls{dnn}-based gravitational model in orbital dynamic analysis.
Instead of focusing on-orbit estimation, \citet{kalita2017network} introduce an \gls{nn} to the formulation of asteroid missions in terms of the planning and design phases, while \citet{feruglio2016neural} utilise a feed-forward \gls{nn} to autonomously identify a \gls{sc} impact event. \citet{viavattene2019artificial} take advantage of a \gls{nn} to map the transfer time and cost for \gls{nea} rendezvous trajectory. 

\acrconnect{dnn}{-based} optical navigation is an increasingly important area in asteroid exploration missions, which can manage challenges of previous schemes, including traditional high-cost and high-risk spacecraft systems, irregular and illuminated asteroids, and conventional image processing techniques. 
In such a scenario (\figref{fig:intro-scenarios-ap}), the chaser may be commanded to inspect a particular patch on the surface of the asteroid it has rendezvoused with (observed on frame $\Fc$), which is intrinsically a localisation task requiring the estimation of $\mT_{ct}$. If the asteroid has been previously mapped, and there exists a codebook with annotated landmarks (on frame $\Ft$) for comparison, there are two possible approaches. The first follows the same direct classification procedure as \citref{campbell2017deep}, where a \gls{dnn} is used to match the observed patch with the corresponding patch in the codebook, which is annotated with the relative pose, but with the dataset of Lunar surface. The alternative approach is to have a single class per patch on the database and train the \gls{dnn} to be robust to viewpoint distortion, and then rely on classical image processing techniques to infer the pose based on the different observed features between the observations and matched patches.

\begin{figure}[!t]
\centering
	\begin{subfigure}[t]{0.99\textwidth}	\centering
		\includegraphics[width=\textwidth]{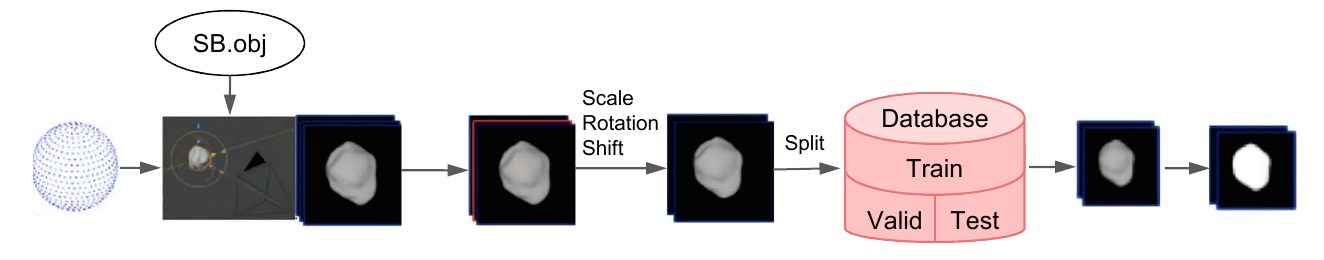}
		\caption{Step flows for database generation}
	\end{subfigure}
	\begin{subfigure}[b]{0.99\textwidth}    \centering
		\includegraphics[width=\textwidth]{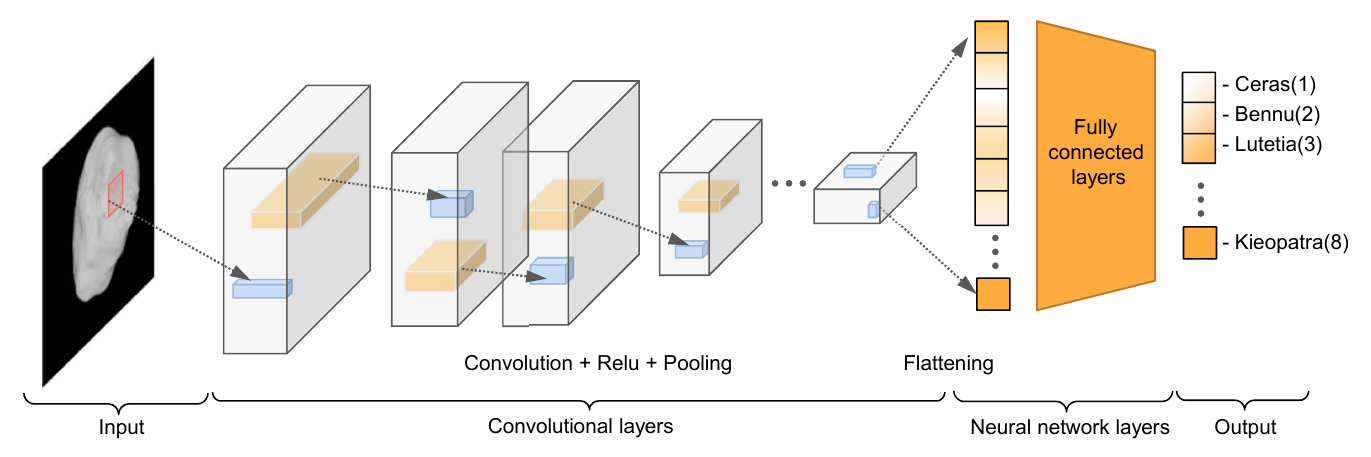}
		\caption{Schematic representation of the proposed \gls{cnn}}
	\end{subfigure}
	\caption{Database generation flow and proposed \gls{cnn} architecture of \citet{pugliatti2020small}.}
	\label{fig:pugliatti2021navigation}
\end{figure}

\citet{pugliatti2020small} first present \acrconnect{cnn}{-based} methods for on-board small-body shape classification since shape information can enhance the image processing and autonomy of self-task planning. A set of \num{8} well-known models from the \gls{pds} node\footnote{\url{https://sbn.psi.edu/pds/shape-models}.} is selected to represent the most important features of small asteroids at a global scale. \figref{fig:pugliatti2021navigation} presents a sketch of the steps for building the database and the proposed \gls{cnn} framework for classifying asteroids. The database is generated in Blender with an assumed camera pointing and illumination, and further augmented in TensorFlow \citep{abadi2016tensorflow} with random rotations, translations, and scaling. The database composed of \num{20988} images is divided into training, validation and test sets according to a \SI{80}{\percent}-\SI{10}{\percent}-\SI{10}{\percent} split. Their \gls{cnn} architecture has five \glspl{cl} in sequence, with each followed by a pooling layer, a reshape operation, and three \glspl{fl} to classify. A hyperparameter search is used to obtain network parameters. Three traditional approaches, such as Hu invariant moments, Fourier descriptors, and polar outlines, are compared, in which the proposed \acrconnect{cnn}{-based} scheme performs best.

Later, \citet{pugliatti2021navigation} proposed on-board autonomous navigation using segmentation maps and a \gls{cnn} to estimate spacecraft position concerning an asteroid fixed reference frame. The \gls{cnn} transferred from the MobileNetV2 network \citep{sandler2018mobilenetv2} classifies the segmentation maps to generate a rough estimate of the position information from the input. The relative position is finally obtained by refining the output of the \gls{cnn} using an advanced normalised cross-correlation method. Didymos and Hartley are selected as representatives of regular and irregular small-bodies to create the dataset, which includes \num{49716} samples of synthetic maps for five different scenarios.  Experimental results indicate the capability of \gls{cnn} in predicting the correct class and achieve a relative position error below \SIrange{5}{8}{\percent} of the range from the asteroid.

In 2021, \citet{ravani2021site} developed a novel Mask-Region \gls{cnn} to detect landing sites for autonomous soft-landing on asteroids. Since there is no open public dataset of potential landing sites labelled with ground truth, the authors first gather image mosaics of the asteroid Vesta from the \gls{pds} of \gls{nasa}\footnote{\url{https://pds.nasa.gov}.} and then fragment the large images into smaller ones with a fixed size. Next, the training dataset is labelled manually. For the Mask‑Region \gls{cnn} pipeline, it follows a backbone network of ResNet-50-C4 for initialisation; an \gls{rpn} for extracting feature maps; Faster \gls{rcnn} for \gls{roi} alignment; the network head is structured using \glspl{fl} for computing the bounding box; and the mask head of a \gls{fl} network \citep{long2015fully} is used for calculating the pixel-level mask. Comparing with conventional image processing methods, the proposed network on their dataset results in an accuracy of \SI{94}{\percent} with lower computational time cost in the implementation phase.

\section{Summary and Conclusion}
\label{sec:conclusion}

\noindent This work surveyed recent trends in deep learning techniques for 6-\gls{dof} relative pose estimation in spaceborne applications. Contributions in the field of computer vision were presented, followed by concrete applications from the literature to autonomous spacecraft navigation, including spaceborne pose estimation, crater and hazard detection of terrain relative navigation, and \gls{dl}-based asteroid navigation. This survey is motivated by the applicability of \gls{dl} techniques in relative spacecraft navigation for future space missions, i.e.\ rendezvous, docking, formation flying, descent and landing on the lunar surface, orbiting and inspecting asteroids. The general \gls{dnn} framework for the applications in this research area was reviewed in terms of network structure, type of network, training method, dataset topology and generation, and attained performance.

First, a review of \gls{dnn}-based \gls{sc} relative pose estimation techniques was given, in which a top level distinction between supervised and unsupervised methods was made, whereby contributions in the space domain were found to belong exclusively to the former. Context in terms of preceding ground-based applications was established. Further lower level categorisations were made; in particular, it was found that many techniques favoured a direct approach (so called ``end-to-end''), where a \gls{dnn} pipeline is trained directly on images to yield the relative state. Indeed, this is a very appealing property of deep learning, as not only is the feature extraction task relayed to a \gls{cnn}, but so is the modelling task, eliminating the ''middleman'' and allowing the user to focus mainly on the architecture design and optimisation of learnable parameters. However, it was seen that more accurate solutions were obtained by combining them with classical methods. For these indirect methods, a \gls{cnn} was tasked with regressing the locations of \gls{2d} keypoints on the target and estimating the relative pose from geometrical correspondences with their \gls{3d} counterparts, using techniques such as \gls{pnp} or nonlinear optimisation. Furthermore, such solutions are easily incorporated into navigation filters to further refine the estimate with continuous, smooth consistency (also beyond pose estimation). The role of \glspl{rnn}, particularly \glspl{lstm}, is highlighted in the processing of a continuous stream of images. Tables \ref{Table:summary-direct} and \ref{Table:summary-indirect} summarises these findings in terms of relative pose estimation error for spacecraft rendezvous \gls{dl} applications. 

Second, the applications of \glspl{dnn} to \gls{trn} were divided into three aspects for surveying, in which the \gls{dnn}-based crater and hazard detection methods were recognised as contributors towards building a terrain navigation system. It was pointed out that public open data for training and testing of \gls{dnn}-based \gls{trn} frameworks is limited. Furthermore, \gls{dl}-based relative navigation methods focusing on asteroid missions were provided. The challenges and motivations were discussed before a detailed review of this field.

Lastly, regarding unsupervised learning methods (i.e.\ concerning cases in which the desired output for each input is not given during training), far too little attention has been paid to this kind of technique for space navigation. However, unsupervised techniques such as \gls{cnn}-\glsxtrshort{slam} (\glsxtrlong*{slam}) or unsupervised \gls{vo} are underlined as a potential novel approach for the space domain and may be investigated in future. \textcolor{black}{Additionally, most publications study the application of \gls{dl} in space in a theoretical way without being concerned with computational performance; indeed, only a few publications \cite{sonawani2020assistive, oestreich2020onorbit, moghe2020online} focus on actual deployments on hardware, considering things like execution time, and size of the training dataset. Therefore, it can be concluded that these studies towards the actual engineering practice have been little discussed and require further development.}
\bibliography{references}

\end{document}